\documentclass{article}

\PassOptionsToPackage{numbers, compress}{natbib}

\usepackage[preprint]{neurips_2023}
\usepackage[utf8]{inputenc} 
\usepackage[T1]{fontenc}    
\usepackage{hyperref}       
\usepackage{url}            
\usepackage{booktabs}       
\usepackage{amsfonts}       
\usepackage{nicefrac}       
\usepackage{microtype}      
\usepackage{xcolor}         
\usepackage{algpseudocode}
\usepackage{changepage}
\usepackage{graphicx}
\usepackage{xspace}

\usepackage{amsmath}
\usepackage{amssymb}
\usepackage{bm}
\usepackage{amsthm}
\usepackage{amssymb}
\usepackage{caption}
\usepackage[ruled,vlined]{algorithm2e}

\usepackage{subcaption}
\usepackage{graphicx}
\usepackage{floatrow}
\usepackage{wrapfig}
\usepackage{lipsum}

\theoremstyle{plain}
\newtheorem{theorem}{Theorem}[section]

\newtheorem{corollary}[theorem]{Corollary}
\theoremstyle{definition}

\newcommand{\defeq}{\mathrel{\mathop:}=}

\title{Diffusion-Jump GNNs:\\
Homophiliation via Learnable Metric Filters}

\author{%
  Ahmed Begga Hachlafi \\
  University of Alicante, Spain\\
  \texttt{ahmedbegga@gmail.com} \\
   \And
   Francisco Escolano \\
   University of Alicante, Spain\\
   \texttt{escolano.ua@gmail.com} \\
   \And
   Miguel Angel Lozano \\
   University of Alicante, Spain\\
   \texttt{malozano@ua.es} \\
   \And
   Edwin R. Hancock \\
   University of York, U.K.\\
   \texttt{edwin.hancock@york.ac.uk} \\
}

\begin{document}

\maketitle

\begin{abstract}
  High-order Graph Neural Networks (HO-GNNs) have been developed to infer consistent 
  latent spaces in the heterophilic regime, where the label distribution is not correlated 
  with the graph structure. However, most of the existing HO-GNNs are \emph{hop-based}, i.e., 
  they rely on the powers of the transition matrix. 
  As a result, these architectures are not fully reactive to the classification loss and the 
  achieved structural filters have static supports. In other words, neither the filters' supports 
  nor their coefficients can be learned with these networks. They are confined, instead, to learn combinations 
  of filters. 
  To address the above concerns, we propose \textsc{Diffusion-jump GNNs}$-$ a method relying on 
  asymptotic diffusion distances that operates on \emph{jumps}. A \emph{diffusion-pump} generates 
  pairwise distances whose projections determine both the support and coefficients of each structural 
  filter. These filters are called jumps because they explore a wide range of scales in order to find  
  bonds between scattered nodes with the same label. Actually, the full process is controlled by the 
  classification loss. Both the jumps and the diffusion distances react to classification errors (i.e. 
  they are learnable). 
  \emph{Homophiliation}, i.e., the process of learning piecewise smooth latent spaces in the heterophilic 
  regime, is formulated as a Dirichlet problem: the known labels determine the border nodes and the diffusion-pump 
  ensures a minimal deviation of the semi-supervised grouping from a canonical unsupervised grouping. This triggers the update of both the 
  diffusion distances and, consequently, the jumps in order to minimize the classification error. 
  The Dirichlet formulation has several advantages. It leads to the definition of \emph{structural heterophily}, a novel 
  measure beyond edge heterophily. It also allows us to investigate links with (learnable) diffusion distances, absorbing random walks and 
  stochastic diffusion. 
  Finally, our experimental results outperform significantly those of the state-of-the-art both in homophilic and heterophilic 
  datasets. We are very competitive for large graphs. 
\end{abstract}

\section{Introduction}\label{sec:1}
The success of Graph Neural Networks (GNNs) relies on their convolutional architecture~\citep{kipf2017semi}\citep{hamilton2017inductive}\citep{velickovic2018gat}. Their  \emph{aggregate and combine} mechanism provides a significant degree of expressiveness. However, in the heterophilic regime, such a mechanism (initially designed for homophilic graphs) results in the over-smoothing issue (shadowing of the internal representations of the nodes, due to a non-selective aggregation). In ~\citep{Beyond2020}, three solutions are explored: (i) ego and neighbor embedding separation, (ii) higher-order neighborhoods, and (iii) a combination of intermediate representations. The purpose of these mechanisms is to enforce the internal representations of the node  features so that the resulting latent space becomes consistent (piecewise smooth), for instance, when the downstream task is node-classification~\citep{Non-Local22}\citep{zhu2021graph}\citep{pmlr-v119-chen20v}. 

In this paper, we explore High-Order GNNs (HO-GNNs). One type of HO-GNNs results from \emph{rewiring the edges} in the graph. For instance, the method in~\citep{bi2022make} explores neighborhoods of several orders (hops) selecting those orders who provide a high correlation between the node features. GATs~\citep{velickovic2018gat} are also a well-known rewiring method: the strength of each edge in the input graph is given by a trainable weight. Such a weight is corrected if the concatenation of the node features associated with the corresponding edge has a negative impact on node-classification. Diffwire~\citep{arnaiz2022diffwire} is another trainable rewiring method. The basic idea of Diffwire is to estimate the commute-times distance between each pair of nodes and use the distance matrix to mask the original adjacency matrix. Other non-differentiable rewiring methods are mainly addressed to alleviate the over-squashing issue (bottlenecks obstruct the message-passing process). A couple of recent examples are \citep{topping2022understanding} and \citep{digiovanni2023oversquashing}. 

A second type of HO-GNNs are \emph{Deep/Sequential hop-based} methods, i.e. those models that address over-smoothing with a deep architecture. GGCNs~\citep{Twosides22} attenuate over-smoothing by performing edge correction (corrected edge weights are learned from node degrees, and signed edges are learned from node features). However, Shortest-Paths-MPNNs~\citep{abboud2022shortest} and Ordered-GNNs~\citep{song2023ordered} are more focused on performing robust aggregations. Shortest-Paths-MPNNs compute the shortest paths between any
pair of nodes. Then, for each node, several separate aggregations are performed (each one for increasing lengths of the shortest paths); then, the resulting embeddings are weighted. Ordered-GNNs rely on a similar principle: for each node, the hierarchy of a tree rooted in that node is aligned with the hops wrt this node in the graph. As neighboring nodes within $k$ hops form a depth$-k$ subtree, aggregations for shallow sub-trees precede those for deeper ones. Interestingly, Ordered-GNNs introduce a differentiable way of deciding the split point between sub-trees.

Finally, \emph{Shallow/Parallel hop-based} methods explore several hop orders in parallel and then integrate the resulting embedding (e.g. via concatenation). 
MixHop~\citep{mixhop19}, FSGNNs~\citep{FSGNN21} and DualNets~\citep{DualNet22} compute several powers $\mathbf{P}^k,\; k=1,2,\ldots, K$
of the normalized adjacency matrix (transition matrix) $\mathbf{P}=\mathbf{D^{-1}A}$. Each power feeds a different 
GNN. The resulting embeddings are weighed and concatenated for later discrimination. SIGN~\citep{sign_icml_grl2020} is similar to 
MixHop but it precomputes the aggregations $\mathbf{P}^k\mathbf{X}$ for the sake of scalability. More recently, the Simple Graph Convolution (SGG) method~\citep{chanpuriya2022simplified} improves MixHop by learning polynomials of the transition matrix.  

Finally, Generalized PageRank GNNs (GPR-GNNs)\citep{GPR21}learns jointly the best embedding of each node feature and the best weight of each hop. This is very interesting, since the suitability of $\mathbf{P}^k$, which is encoded by a weight $\gamma_k$, influences the latent space of the features $\mathbf{H}^0=f_{\Theta}(\mathbf{X})$ through a learnable function $f_{\Theta}(.)$. As a result the $k-$th embedding is  $\mathbf{H}^k = \mathbf{P}^k\mathbf{H}^0$. This mechanism allows GPR-GNNs to avoid over-smoothing and trade node and topology feature informativeness. However, this strategy produces inconsistent results since GPR-GNNs are better suited for heterophilic graphs, instead of being also useful for homophilic graphs. 


\begin{wrapfigure}{r}{0.35\textwidth}
    \centering
    \includegraphics[width=0.6\textwidth,height=5cm]{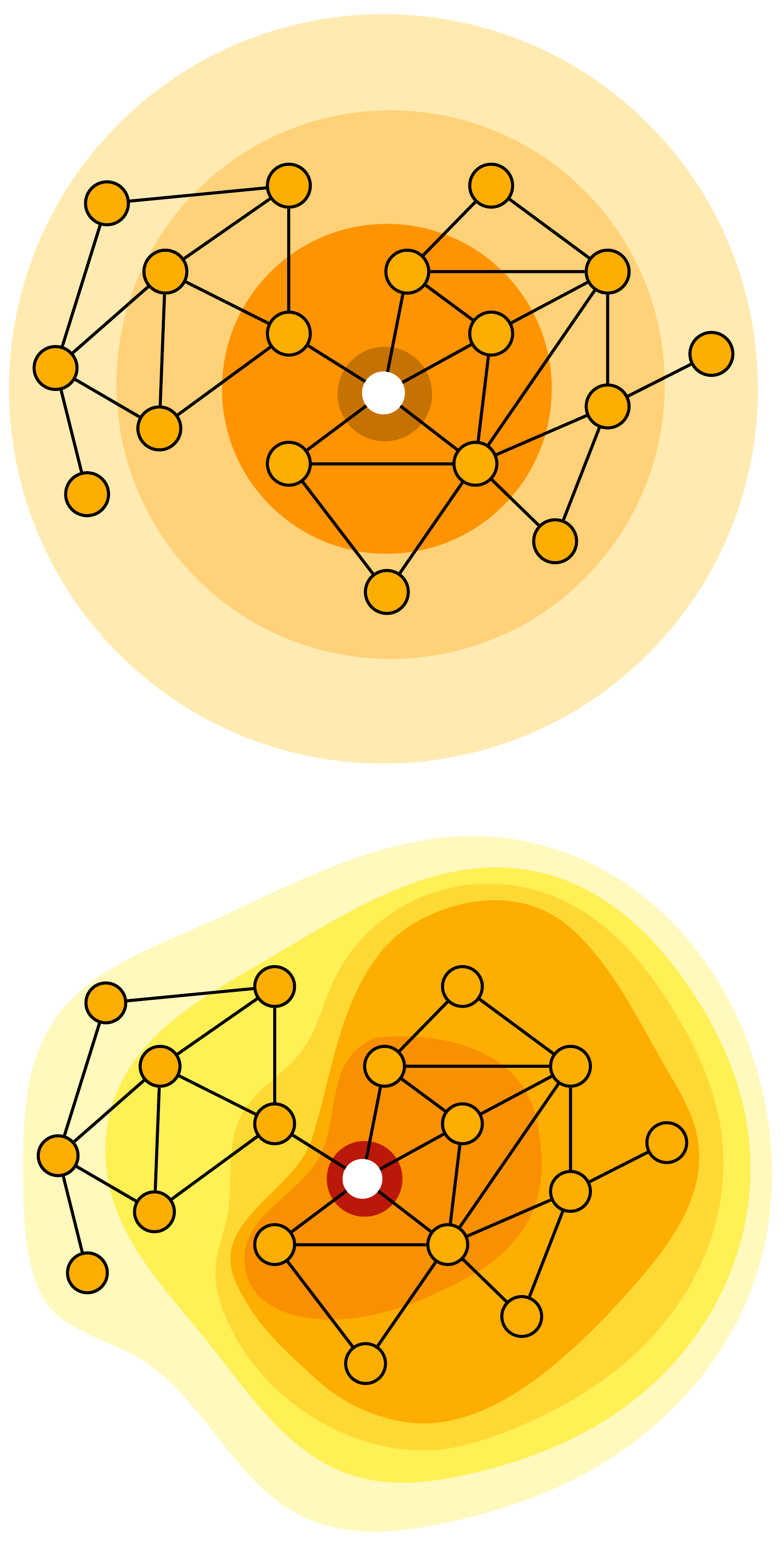}
    \caption{Hop-hierarchy (Top) vs Jump-hierarchy (Bottom). Diffusion distances contract the similarity space due to \emph{structural forces}.}
    \label{fig:jump}
\end{wrapfigure}

\textbf{Main Limitation of HO-GNNs.} Most of the existing HO-GNNs explore different powers 
of the normalized adjacency matrix (transition matrix) $\mathbf{P}$. In  other words, they are completely \emph{hop-based}. 
As a result, the HO-GNNs exploit the labels of the semi-supervised learning process either to alleviate the over-smoothing issue (in the sequential case~\citep{Twosides22}\citep{abboud2022shortest}~\citep{song2023ordered}) or to weigh the importance of each hop order (in the parallel case~\citep{mixhop19}\citep{FSGNN21}\citep{DualNet22}\citep{sign_icml_grl2020}). However, as the structure of the input graph is \emph{static}, the hops are static as well. Consequently, \emph{the labels cannot be backpropagated 
to change the structure of the hops, but only the relative importance of each hop or the extent of its  aggregation support}.

\textbf{Implications.} As a result, dealing with heterophilic graphs goes beyond the potential achievements of hop-based approaches (see our Experiments in Section~\ref{sec:5}). Despite high-order hops being able of connecting distant 
nodes with the same label, such connections can be neither attenuated nor amplified for the sake of the classification loss. In other words, the probability that a random walk links two nodes is an \emph{in-place coefficient}, not the realization of a probabilistic event. In this regard, parallel HO-GNNs claim that the powers $\mathbf{P}^k$ can be interpreted as a bank of \emph{structural filters}, i.e. a bunch of aggregators inspired by convolutional filters such as Gabor receptive filds~\citep{mixhop19}. However, an expressive characterization of a structural filter requires that both its \emph{support} (specification of what edges have a non-zero coefficient) and its \emph{coefficients} are learnable.  

\textbf{Our contributions.} In this paper, we address the problem of learning a bank of expressive structural filters as follows:
\begin{enumerate}
    \item[\textbf{a)}] We re-formulate the problem of node-classification under heterophily in terms of a \textbf{Dirichlet problem}, i.e. we have \emph{border nodes} (training set) where the classification is optimal and \emph{interior nodes} (remaining nodes) where the resulting latent space (node embedding) must be as harmonic (piecewise smooth) as possible, even when the labeling is far from being harmonic over the graph. 
    \item[\textbf{b)}] We have a \textbf{diffusion pump} which generates \emph{asymptotic diffusion distances} $d(i,j) = d^{t\rightarrow\infty}(i,j)$ between the nodes by learning the nontrivial top eigenvectors of $\mathbf{P}$ subject to the labeling of the training set. 
    \item[\textbf{c)}] Given the diffusion distances $d(i,j)$ we compute the \emph{jump hierarchy} (see Figure~\ref{fig:jump}). For a node $i$ we have that $i\in J_i^0$ ($k=0$) is the closest node wrt itself; $j_1\in J_i^1$ ($k=1$) are nodes so that only $j_1$ is closer to node $i$ than any of them, $j_2\in J_i^2$ ($k=2$) are nodes so that only the nodes in $J^1$ are closer to $i$ than any of them, and so on. Each of the sets $J^k = \bigcup_{i=1}^ {|V|}J_i^k$, where $V$ are the nodes of the graph $G=(V,E)$, is called a \textbf{jump}. 
    \item[\textbf{d)}] The edges $E_k=\{(i_k,j_k)\in V\times V:i_k,j_k\in J^k\}$ define the \textbf{support of the jump} and the \textbf{coefficients} $c(i_k,j_k)=g(d(i_k,i_k))$ are given by a function $g(.)$ of the diffusion distances (for example the neg-exponential). Then, the \textbf{structural filter} $\mathbf{J}^k$ is a matrix with non-zero coefficients only at $E_k$. 
    \item[\textbf{e)}] Each structural filter $\mathbf{J}^k$ with $k=0,1,\ldots, K$ feeds a GNN parameterized by $\mathbf{W}^k$ and the resulting embedding  $\mathbf{H}^k=\sigma(\mathbf{J}^k\mathbf{X}\mathbf{W}^k)$ is weighted by a learnable parameter $\alpha_k$ subject to  $\sum_{k=0}^K\alpha_k=1$. All the weighted embeddings are concatenated and feed a forward network for classification. 
\end{enumerate}

In addition, we propose a novel metric called \textbf{structural heterophily} and we denote the process of generating homophilic embeddings from heterophilic graphs as \textbf{homophiliation}.

This paper is organized as follows. In Section~\ref{sec:2}, we formulate semi-supervised learning in terms of a Dirichlet problem. This allows us to measure heterophily in a structural way. Section~\ref{sec:3} is devoted to formulating the loss functions and explaining the dynamics of the optimization process. This is done through the analysis of the main modules of our model. Section~\ref{sec:4} provides more technical and formal details of our model and establishes links with related inspiring formulations. Our experiments are presented and discussed in Section~\ref{sec:5}. Finally, our conclusions and future work are summarized in Section~\ref{sec:6}.

\section{Heterophily as the Loss of Harmonicity}\label{sec:2}\textbf{Node-classification under heterophily} can be posed as the following semi-supervised learning problem. Given an input graph, $G=(V, E)$ with adjacency matrix $\mathbf{A}$ and node features $\mathbf{X}$, there is a node subset $B\subset V$ whose labels $\ell(B)$ are known by the learner (border nodes). Similarly, the labels $\ell(U)$ of the remaining nodes, those in $U = V\sim B$, are hidden (unknown nodes). 

Given the graph Laplacian $\triangle$, and a regularizer (minimizer of $\mathbf{x}^T\triangle\mathbf{x}\defeq\sum_{i\sim j}(\mathbf{x}_i-\mathbf{x}_j)^2)$, we have $\ell^{\ast}=\arg\min_{\ell} \ell^T\triangle\ell$, where $\ell^{\ast}$ is the smoothest labeling of $V$ after propagating $\ell(B)$ to $\ell(U)$ through the edges of the graph. A Dirichlet solver ensures that the labeling $\ell^{\ast}$ is \emph{Harmonic} (the label of a given unknown node is the average of those of its neighbors) subject to the labeling of the border nodes $\ell(B)$.

In the heterophilic regime, two neighboring nodes rarely share their labels. As a result, ${\ell^{\ast}}^T\triangle\ell^{\ast} \gg \mathbf{u}^T\triangle\mathbf{u}$, where $\mathbf{u}$ are the vectorized labels obtained by an alternative unsupervised learner. The unsupervised learner typically assumes that the labels $\mathbf{u}$ are correlated with the topology of the graph (homophily). In other words, \emph{heterophily can be posed in terms of how much harmonicity is lost wrt the homophilic assumption}.  

The objective of a GNN is to learn a parametric function $f_{\Theta}(\mathbf{A},\mathbf{X},\ell(B))$ returning $\mathbf{H}$, a matrix (embedding) of latent representations (one row per node) so that the embeddings of either border nodes or hidden nodes with the same label are grouped together. However, $f_{\Theta}(.)$ does not necessarily minimize $c(\mathbf{H})^T\triangle c(\mathbf{H})$, where $c(.)$ contains the vectorized classification labels. We need to infer a \emph{hidden graph} $G'=(V,E')$ where $c(\mathbf{H})^T\triangle_{G'} c(\mathbf{H})$ is minimized. Actually, the edges $E$ in the hidden graph should link nodes with the same label, even if they are not in the same community. 

\textbf{Structural Heterophily.} Given the above formulation we may characterize heterophily in a structural way, namely \emph{as the departure from a structural unsupervised grouping}. In particular, the ratio 
\begin{equation}\label{structuralh}
    {\cal R} = \frac{\ell^T\triangle\ell}{\mathbf{u}^T\triangle\mathbf{u}}\ge 1\;, \text{where}\; \ell\; \text{is the ground-truth labeling},
\end{equation}
 is close to the unit if the graph is homophilic (the structure is completely correlated with the labels). For ${\cal R}>1$ the graph is  heterophilic. The larger the ratio the larger the heterophily. 

We use the example in Figure~\ref{fig:jump} to illustrate how ${\cal R}$ works. We have two  communities, $V=A\bigcup\bar{A}$ (left and right respectively). The \emph{white node} belongs naturally to the right one $\bar{A}$, and this is what an unsupervised structural clustering detects: the Fiedler vector $\mathbf{u}=\arg\min_{\mathbf{x}\neq\mathbf{0},\mathbf{x}\perp\mathbf{1}}\mathbf{x}^T\triangle\mathbf{x}$ has positive components ($\approx +1$) in $A$ and negative components ($\approx -1$) in $\bar{A}$. 

The vector $\mathbf{u}$ is the smallest nontrivial eigenvector of $\triangle$ as well as the largest nontrivial eigenvector of $\mathbf{P}$. It has been argued that the top eigenvectors of $\mathbf{P}$ may be used to decompose the state space into metastable subspaces~\citep{meta04}. In other words, each of the two graph communities in Figure~\ref{fig:jump} defines a metastable state from which a random walker tries to escape (Section~\ref{sec:4}). 

The average escape time is the inverse of the top nontrivial eigenvalue of $\mathbf{P}$, i.e. the inverse of the approximate spectral gap~\citep{siam81}\citep{Kramers90}\citep{sclimitations06}. In our example, the spectral gap is very tiny so we can expect large escape times (see more details in Section~\ref{sec:5}). In particular, the two states defined by the Fiedler vector are very compact (they have low variability). As a result, all the pairs of nodes $(i,j)$ inside each community have very similar asymptotic diffusion distances~\citep{difmaps05} $d^{t\rightarrow\infty}(i,j)$ according to the structural forces characterizing each metastable state.
    
Consequently, the jump hierarchy defines a succession of unstable states $\mathbf{u}_1,\mathbf{u}_2,\ldots$ resulting from the expansion from $\bar{A}$: $\bar{A}\subseteq \bar{A}_1\subseteq \bar{A}_2\subseteq\ldots$. They are unstable because their Dirichlet energies $\mathbf{u}_k^T\triangle \mathbf{u}_k$ are greater than that of the unsupervised clustering (\emph{ground energy}) $\mathbf{u}^T\triangle \mathbf{u}$.

Last, but by no means least, if we label the white node as belonging to $A$ instead of belonging to $\bar{A}$ (i.e., \emph{we introduce heterophily}), we also increase the Dirichlet energy wrt the ground energy, i.e. $\ell^T\triangle\ell > \mathbf{u}^T\triangle \mathbf{u}$. Why? This is because the new Fiedler vector $\mathbf{u}_{\ell}$ leading to the labeling $\ell$ does no longer induce a step function but a hyperbolic tangent with a positive slope. This is consistent with the increase of the spectral gap and the reduction of the escape time. 

Therefore, one useful interpretation of heterophily in structural terms (departure from the ground energy) is the fact that heterophily relaxes Dirichlet energies in such a way that it is possible to escape from a community in a few jumps and then find nodes with the same label in other communities. Therefore, paying attention to several jumps simultaneously increases the chance of aggregating distant nodes with the same label, thus solving the heterophily issue.

\begin{figure*}[ht]
\begin{center}
\centerline{\includegraphics[width = \linewidth,height =7cm]{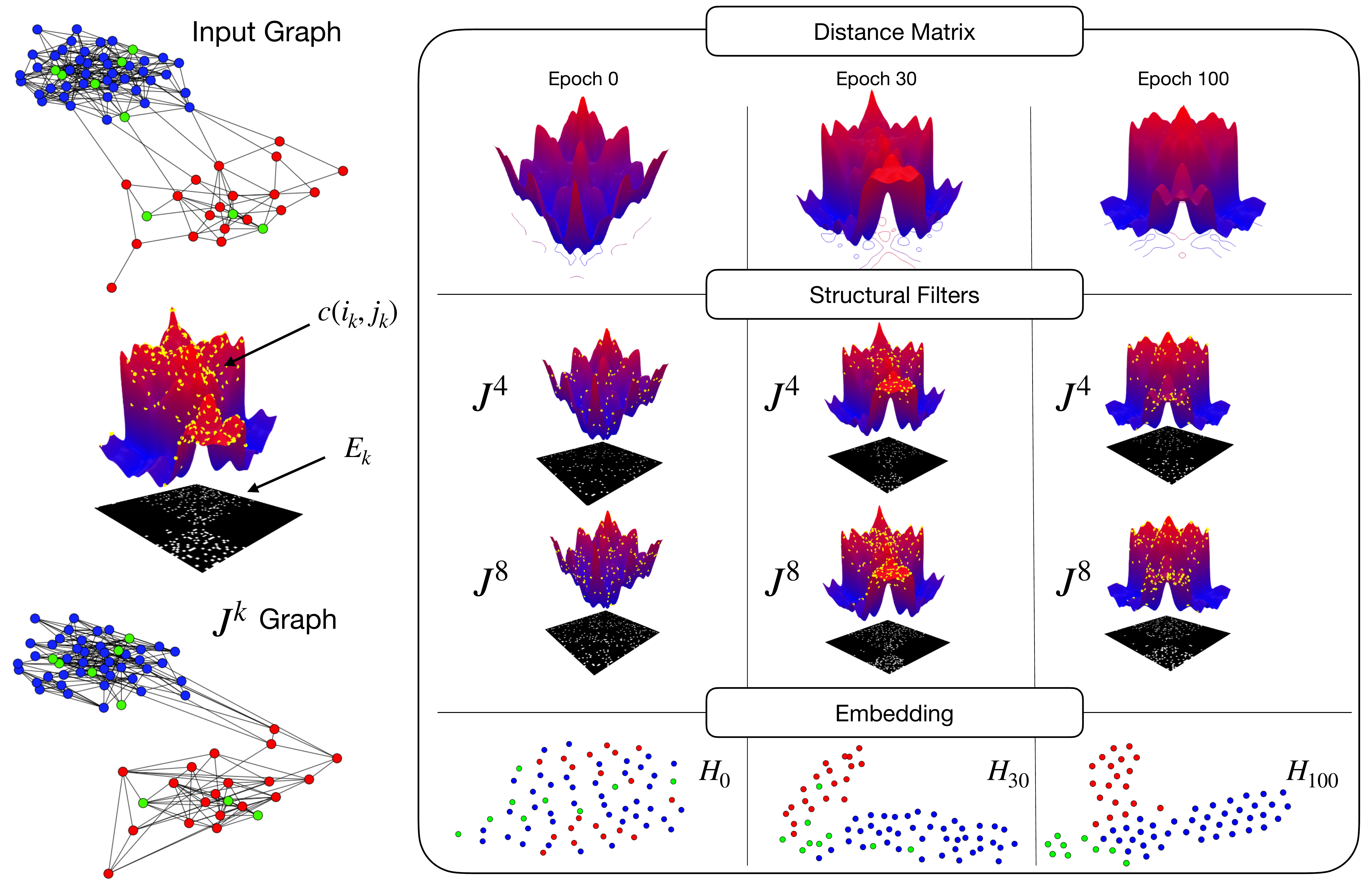}}
    \caption{Homophiliation while the jumps and pairwise distances are learned. Top-Left: a heterophilic graph. Center-Left: The current distance matrix ${\cal D}$ leads to a tridimensional weight distribution $c(i,j)=e^{-d(i,j)}$. Yellow points denote the support $E_k=\{(i_k,j_k)\}$ of the filter $\mathbf{J}^k$. The filter coefficients are given by the weights of the support $C_k=\{c(i_k,j_k)\}$. Bottom-Left: graph used for aggregation with this filter. In the right panel, we show the weight distribution (top), a couple of filters (middle), and the resulting homophiliation (bottom) for some epochs. In particular, we show $\mathbf{H}_0$, $\mathbf{H}_{30}$ and $\mathbf{H}_{100}$. In each epoch $e$, all the embeddings $\{\mathbf{H}_e^k\}$ contribute to identifying potential links between scattered nodes with similar labels. If any of these links is wrong, the matrix of pairwise distances ${\cal D}$ is updated.}.
    \label{fig:jump-explain}
\label{figure-pump}
\end{center}
\end{figure*} 
   
\section{Homophiliation: Losses and Modules}\label{sec:3}

\textbf{Homophiliation.} Our computational model for node-classification under heterophily cannot only rely on finding Harmonic labelings but also on transforming the matrix of node features $\mathbf{X}$ into a piece-wise smooth embedding $\mathbf{H}$. The rows in $\mathbf{H}$ associated with nodes with the same label must be clustered together and these labels must be consistent with those of the border nodes $\ell(B)$. Such a process, i.e. the learning of $f_{\Theta}(\mathbf{A},\mathbf{X},\ell(B))$, results from solving the following optimization problem:
\begin{align}
\label{eq:loss}
\text{Min}\;\;   &  {\cal L} = \text{Tr}[f_{\theta}(\mathbf{A})^T\triangle f_{\theta}(\mathbf{A})] + {\cal L}_{c}(\{\mathbf{J}^k\},\mathbf{X},\ell(B)) \nonumber\\
\text{s.t.}\;\;  & \mathbf{U}^T\mathbf{U}=\mathbf{I}  \nonumber\\
                 &  \mathbf{U}=f_{\theta}(\mathbf{A}),\; {\cal D}(i,j)=\|\nabla \mathbf{U}_{ij}\| \;\; \text{and}\;\; \mathbf{J}^k=\Pi^k\cdot\exp\left(-{\cal D}\right)\;,           
\end{align}
where we have an interplay between the \emph{Dirichlet loss} $\text{Tr}[f_{\theta}(\mathbf{A})^T\triangle f_{\theta}(\mathbf{A})]$ and the \emph{classification loss} ${\cal L}_{c}$(Cross-Entrophy)  as follows. 

\textbf{Diffusion Pump.} Minimizing the structural heterophily so that ${\cal R}\approx 1$ (Eq.~\ref{structuralh}) in $G'=(V,E')$ implies learning Dirichlet energies close to the ground energy. However, in the heterophilic regime, we cannot minimize $c(\mathbf{H})^T\triangle_{G'} c(\mathbf{H})$ before discovering the optimal embedding $\mathbf{H}^{\ast}$. In the meanwhile, the Dirichlet formulation allows us to learn the smallest nontrivial eigenvectors of $\triangle$ as we do in the unsupervised setting (e.g. the Fiedler vector). These eigenvectors will be in the columns of $\mathbf{U}$, but they do not have a free form. The notation $\mathbf{U}=f_{\theta}(\mathbf{A})$ in the above optimization problem goes beyond emphasizing the learnability of $\mathbf{U}$. We also constrain the eigenvectors to be projections/transformations of the adjacency matrix $\mathbf{A}$. 

We learn the eigenvectors $\mathbf{U}$ because it is key to computing diffusion distances between the nodes. In the following, we will replace $d(i,j)$ by ${\cal D}(i,j)$ when we need to emphasize the matrix nature of the pairwise distances. Each pairwise distance ${\cal D}(i,j)$ comes from the norm of $\nabla \mathbf{U}_{ij}=\mathbf{U}_{i:}-\mathbf{U}_{j:}$ (row-difference). As we will detail in Section~\ref{sec:4} , $\|\nabla \mathbf{U}_{ij}\|$ approximates  the asymptotic diffusion distance between two nodes $i$ and $j$. Herein, we focus on the fact that nodes belonging to the same sub-structure (e.g. cluster or community) have similar distances. Back to Figure~\ref{fig:jump}, if $i$ and $j$ belong to the same community, two random walks placed in $i$ and $j$ have similar escape probabilities. Therefore, we build a hierarchy of escape probabilities to characterize the respective reachability of any node wrt a given one. Interestingly, the hierarchy induced by hops is isotropic wrt each node whereas the hierarchy induced by escape probabilities is anisotropic. This latter hierarchy is built by specifying binary projection matrices $\Pi^k$ which select the pairs of distances that support the creation/update each structural filter $\mathbf{J}^k$. 

\textbf{Exploration by Parallel Jumping.} The diffusion pump triggers the creation of $K+1$ structural filters $\{\mathbf{J}^0,\mathbf{J}^1,\ldots,\mathbf{J}^K\}$ derived from their respective jumps $J^k$. Each filter $\mathbf{J}^k$ has its support $E_k=\{(i_k,j_k)\}$ and its coefficients $c(i_k,j_k)= e^{-d(i_k,i_k)}$. We illustrate this process in Figure~\ref{fig:jump-explain} over a heterophilic graph. At any epoch, the optimizer creates a distance matrix ${\cal D}$ and weighs it: $C=\exp[-{\cal D}]$. The result is a weight distribution (middle-left). Each yellow point in the weight distribution belongs to the jump $J^k$. Consequently, the yellow points denote the edges of the filter support $E_k$ and they are projected in the adjacency matrix below the distribution. The coefficients $c(i_k,j_k)$ of the filter are the heights of the yellow points. Finally, the edges in the graph depicted below the distribution are exactly those of the filter support. Neighbor aggregation wrt this filter $\mathbf{J}^k\mathbf{X}$ will be constrained to that graph and these weights. 

The right panel in Figure~\ref{fig:jump-explain} shows the evolution of the weight distribution (top), some filters (center), and the status of the homophiliation process (bottom). The optimization process is initially dominated by the diffusion pump since random weight distributions are explored first. As a result, scattered nodes with similar labels can be potentially aggregated: we start to implicitly build the hidden graph $G'=(V,E')$. The probability of aggregating distant nodes is leveraged by the fact that, during the first epochs, most of the $K+1$ filters $\mathbf{J}^k$ have a random nature independently of  $k$, the filter order. Escape probabilities are relaxed during this \emph{exploration stage}.  

\textbf{Classification Loss.} Each filter, $\mathbf{J}^k$ becomes the aggregator of a naive GNN $\sigma(\mathbf{J}^k\mathbf{X}\mathbf{W}^k)$ which generates an embedding $\mathbf{H}^k$. This embedding is weighted by a learnable parameter $\alpha_k$ and concatenated with the remaining embeddings to feed a classification layer. Therefore, as soon as the structural filters discover interesting bonds for minimizing ${\cal L}_c$, the weights $\mathbf{W}^k$ of all the GNNs, the filters' coefficients, and the distance matrix will become more and more stable. At some point in the optimization process, the Dirichlet loss will be stabilized and the exploration stage ends. Later on, the classification loss will refine the almost-homophilic global embedding $\mathbf{H}$.
As a result, the embeddings of either border nodes or hidden nodes with the same label are grouped together in the latent space (for instance, see the column of Epoch $100$ in Figure~\ref{fig:jump-explain}).





\section{Methodological Details}\label{sec:4}
\subsection{Network Architecture}
\textsc{Diffusion-Jump GNNs} are neural networks $f_{\Theta}(\mathbf{A},\mathbf{X},\ell(B))$ resulting from the optimization problem stated in Eq.~\ref{eq:loss}.  
The interplay between the Dirichlet loss and the classification loss is described above. In this Section,  we give some technical details about the architecture of the network. 

\textbf{Diffusion pump.} The pump is responsible for generating and updating the matrix of pairwise diffusion distances ${\cal D}$. For the generation, we solve any of the following equivalent problems: 
\begin{equation}\label{eq:maxminlosses}
\text{Min}\;\;\frac{Tr[\mathbf{U}^T\triangle\mathbf{U}^T]}{Tr[\mathbf{U}^T\mathbf{D}\mathbf{U}^T]}\; \equiv\;
\text{Max}\;\;\frac{Tr[\mathbf{U}^T\mathbf{A}\mathbf{U}^T]}{Tr[\mathbf{U}^T\mathbf{D}\mathbf{U}^T]}\;,
\end{equation}
both s.t. $\mathbf{U}^T\mathbf{U}=\mathbf{I}$, where $\mathbf{U}_{n\times p}=f_{\theta}(\mathbf{A})$, $n=|V|$. Since $\mathbf{D}$ is the diagonal degree matrix, we have $\triangle\defeq \mathbf{D}-\mathbf{A}$. 
As a result, the Min problem \emph{approximates} the $p$ smallest nontrivial eigenvectors of the normalized Laplacian $\tilde{\triangle}\defeq \mathbf{D}^{-1/2}\triangle \mathbf{D}^{-1/2}=\mathbf{I}-\tilde{\mathbf{A}}$, where $\tilde{\mathbf{A}}\defeq \mathbf{D}^{-1/2}\mathbf{A}\mathbf{D}^{-1/2}$ is the normalized adjacency. Equivalently, the Max problem \emph{approximates} the $p$ largest nontrivial
eigenvectors of the transition matrix $\mathbf{P}\defeq \mathbf{D}^{-1}\mathbf{A}$. Note that $\tilde{\mathbf{A}}$ and $\tilde{\mathbf{P}}$ have the same eigenvectors and also that if $\lambda$ is an eigenvalue of $\mathbf{P}$ then $1-\lambda$ an eigenvalue for $\tilde{\triangle}$. Note also, that we use "\emph{approximates}" instead of "\emph{finds}". This is due to the limitations of Stochastic Gradient Descent (SGD) when solving the Trace-Ratio problems~\citep{TraceRatio07}\citep{TraceRatio09}\citep{TraceRatio12} in Eq.~\ref{eq:maxminlosses}. In this regard, we have the following results with practical implications:

\begin{theorem}[Fiedler Environments]\label{th:1} The SGD solution of the Trace-Ratio Min problem in Eq.~\ref{eq:maxminlosses} can be posed in terms of $\text{Min}\; \text{Tr}[\mathbf{U}^T(\triangle - \rho\mathbf{D})\mathbf{U}]$ under orthonormality constraints. This leads to $\triangle\mathbf{U}=\rho^{\ast}\mathbf{D}\mathbf{U}$, i.e. to the orthogonal eigenfunctions of the normalized Laplacian $\tilde{\triangle}$ associated with $\rho^{\ast}$. However, $\rho^{\ast}$ is not necessarily an eigenvalue of $\tilde{\triangle}$, but an approximation of the Fiedler value $\lambda_2$: $\exists\; \epsilon>0\;:|\lambda_2 - \rho^{\ast}|<\epsilon$. As a result, the $p$ columns $\mathbf{u}_i$ of $\mathbf{U}$ satisfy: $\exists\; \delta>0\;:\|\phi_2 - \mathbf{u}_i \|< \delta$, where $\phi_2$ denotes the Fiedler vector. Then, we obtain what we call a \emph{Fiedler environment}.
\end{theorem} 
\begin{corollary}[Asymptotic Diffusion Distances]\label{cor:1} The norm of $\nabla \mathbf{U}_{ij}\defeq\mathbf{U}_{i:}-\mathbf{U}_{j:}$ (row-difference) is proportional to the approximate commute time between nodes $i$ and $j$, which is $d(i,j)^{t\rightarrow\infty}$, where   $d(i,j)\defeq\|\nabla \mathbf{U}_{ij}\|$. Therefore, the matrix ${\cal D}$ relies on Euclidean distances.  
\end{corollary}
We proof both results in \emph{Appendix A: Formal Results with Practical Implications}. We also give there practical evidence of the need of solving Trace-Ratio problems in an SGD context, instead of solving the original Trace problem in Eq.~\ref{eq:loss}. We also justify the convenience of conditioning $\mathbf{U}$ to $\mathbf{A}$, $\mathbf{U}=f_{\theta}(\mathbf{A})$. Actually, this setting is inspired in how the LINKX method~\citep{LINKX21} exploits the graph topology. 

\textbf{Jumps and Filters}. The bank of learnable structural filters $\{\mathbf{J}^0,\mathbf{J}^1,\ldots,\mathbf{J}^K\}$ is the core of the high-order exploration. Each filter $\mathbf{J}^k$ has a \emph{support} 
$E_k=\{(i_k,j_k)\in V\times V:i_k,j_k\in J^k\}$ with the pairs of nodes (edges of the filter) belonging to the jump $J^k$. In Eq.~\ref{eq:loss}, this is implicitly defined with the expression $\mathbf{J}^k=\Pi^k\cdot\exp\left(-{\cal D}\right)$, where $\Pi^k$ is a $\{0,1\}^{n\times n}$ projection matrix defined, for $k>0$, as follows: 
\begin{equation}\label{eq:pika}
  \Pi^k(i,j) =\left\{\begin{array}{cc}
    1 - \Pi^{k-1}(i,j)  &  \text{if}\; j\in \text{Idx}[\text{top}k^{-1}(i)]\\
    0   &  \text{otherwise}
  \end{array}\right.  
\end{equation}
with $\Pi^0=\mathbf{I}$ and $\text{top}k^{-1}(i)=\{d(i,j_1),\ldots, d(i,j_k)\}$, where $d(i,j_{l})\le d(i,j_{l+1})$ for $j_{l}, j_{l+1}\in V$ and $l=1,2,\ldots,k-1$. 
Then, $\text{Idx}[\text{top}k^{-1}(i)]$ are the sorted positions of the distances wrt the node $i$, i.e. the \emph{distance ranks}. In this way, the product $\mathbf{J}^k=\Pi^k\cdot\exp\left(-{\cal D}\right)$ is derivable wrt ${\cal D}$ as in~\citep{gao2019graph}. Alternatively, we could also rely on the $\text{topK}$ network~\citep{Topk20}.

\textbf{Individual GNNs.} Each structural filter $\mathbf{J}^k$ feeds a vanilla GNN which obtains a partial embedding $\mathbf{H}^k=\sigma(\mathbf{J}^k\mathbf{X}\mathbf{W}^k)$. The GNN also receives the $n\times F$ matrix of node features $\mathbf{X}$, and $\mathbf{x}_i$ denotes the transpose of the $i-$th row of $\mathbf{X}$. Since $\mathbf{J}^k=\Pi^k\cdot\exp\left(-{\cal D}\right)$, then, for a given node $i$, its aggregation is given by $\mathbf{x}_i = \sum_je^{-d(i,j)}\mathbf{x}_j$ instead of being $\mathbf{x}_i = \mathbf{P}^k\mathbf{x}_i$ as in MixHop~\citep{mixhop19} or $\mathbf{x}_i = \left(\sum_k\beta_k\mathbf{P}^k\right)\mathbf{x}_j$ as in Simple Graph Convolution (SGG)~\citep{chanpuriya2022simplified}. As the asymptotic diffusion distances are approximations of commute times (Theorem~\ref{th:1}), our aggregation works as a kernel depending on learnable Euclidean distances. 

\textbf{Combining GNNs.} Each partial embedding $\mathbf{H}^k=\sigma(\mathbf{J}^k\mathbf{X}\mathbf{W}^k)$ is weighted by a learnable parameter $\alpha_k$, where all the $\alpha_k$ form a convex combination. Then, we concatenate all the weighted embeddings to form the global embedding $\mathbf{H}\defeq {\|}_{k=1}^K \alpha_k \mathbf{H}^k = {\|}_{k=1}^K\alpha_k \sigma\left(\mathbf{J}^k\mathbf{X}\mathbf{W}^k \right)$. Since $\mathbf{H}$ feeds an MLP in order to minimize the classification loss ${\cal L}_c$ as in MixHop, the global embedding tends to retain the best partial embeddings for each node. 

\textbf{Homophilic Branch.} One limitation of our method is that setting a small value for the hyperparameter $K$ is not enough to deal with homophilic graphs. For this reason, we have added an extra GNN (the homophilic branch) that works as follows: $\mathbf{H}^{HB}=\sigma(\mathbf{A}\mathbf{X}\mathbf{W}^{HB})$. Therefore we concatenate $\mathbf{H}\defeq\mathbf{H}|\alpha_{HB}\mathbf{H}^{HB}$, where $\sum_{k=0}^K\alpha_k + \alpha_{HB}=1$. See the optimal learned coefficients in Figure~\ref{fig:alfas}.




\subsection{Inspiring Methods}
We conclude this Section by reviewing some \textbf{links with very inspiring methods} in the literature. For instance, our Dirichlet formulation is inspired by classical graph-based semi-supervised methods. In particular, the work in~\citep{Zhou03} addresses the problem of propagating known labels $\ell(B)$ to unknown nodes $u\in U$. Let $\mathbf{Y}$ be a $n\times c$ matrix where $\mathbf{Y}(i,j)=1$ means that node $i\in B$ has label $j$ and $\mathbf{Y}(i,j)=0$ otherwise. Then, we have the following result:
\begin{theorem}[Dichilet label propagation~\citep{Zhou03}]\label{th:2}
    The optimal label of each node $i$ is given by $\mathbf{y}(i)=\arg\max_{j\leq c}\mathbf{F}(i,j)$, where $\mathbf{F}=\beta\left(\mathbf{I}-\alpha\mathbf{P}\right)^{-1}\mathbf{Y}$, being $\mathbf{P}$ the transition matrix and $\alpha + \beta = 1$. In addition, $\mathbf{F}$ minimizes $\frac{1}{2}\left(\text{Tr}[\mathbf{F}^T\triangle \mathbf{F}] + \mu\sum_i\|\mathbf{F}_{i:} - \mathbf{Y}_{i:}\|^2\right)$, where $\mu>0$ is a regularization parameter satisfying $\alpha=1/(1+\mu)$.
\end{theorem}
Consequently, the diffusion pump in our model is governed by a similar equation: Eq.~\ref{eq:loss}. We sketch the proof of the above theorem and its relationship with absorbing random walks~\citep{doyle2000random} and semi-supervised image segmentation~\citep{Grady06} in \emph{Appendix A}. 

Finally, another important source of inspiration was the design of escape probabilities in terms of diffusion equations. Actually, there is a substantial body of theory linking spectral clustering, random walks, diffusion distances, and metastable states~\citep{pmlr-vR3-meila01a}\citep{difmaps05}\citep{sclimitations06} to be analyzed also in the same appendix. Herein, we only highlight the following result: 
\begin{theorem}[~\citep{sclimitations06}]\label{th:3}
Given a probability function in Boltzmann form $p(\mathbf{x})=e^{-U(\mathbf{x})}$ in a given latent space $\mathbf{X}$, the random walk with transition matrix $\mathbf{P}$ converges to the stochastic differential equation $\dot{\mathbf{x}}(t) = - \nabla U(\mathbf{x}) + \sqrt{2}\dot{\mathbf{w}}(t)$, where $\mathbf{w}$ denotes Brownian motion. Also, the potential time scales describing the expected time of passage between clusters rely on the potential function $U(\mathbf{x})$.
\end{theorem}
\section{Experiments and Discussion}\label{sec:5}
\label{discussion}
\begin{wrapfigure}{r}{0.5\textwidth}
    \centering
    \includegraphics[width=\textwidth,height=4cm]{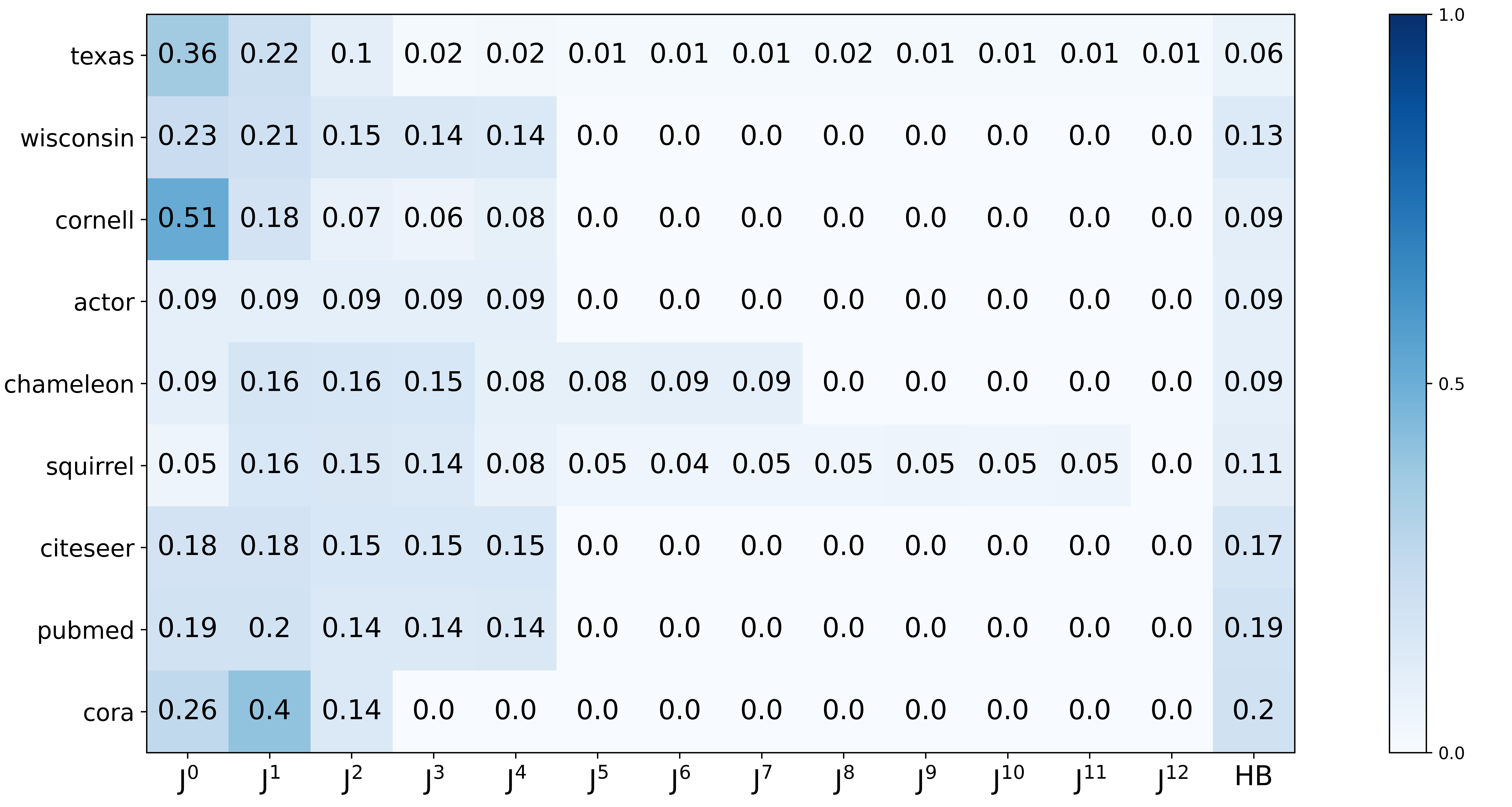}
    \caption{Optimal attention for each jump.}
    \label{fig:alfas}
\end{wrapfigure}
\textbf{Experimental settings}.Table~\ref{tab:results} presents the results obtained by each model on the standard small-medium datasets~\citep{Cora}~\citep{GEOM-GCN}~\citep{actor}. To ensure consistency, we used the same 10 random splits (48\%/32\%/20\%) provided by~\citep{GEOM-GCN}, along with the best configuration for each model. We place the code at \url{https://anonymous.4open.science/r/Diffusion-Jump-GNNs-8EE2/}. The above configurations were extracted from Table 1~\citep{GRAFF} and Tables 3, 10, and 11 in~\citep{LINKX21}. Overall, our model outperformed all others or achieved a close second place, demonstrating strong competitiveness. We assessed the degree of \emph{structural heterophily} using our metric ${\cal R}$. Specifically, we examined two heterophilic regimes: ${\cal R}<8$ indicating low structural heterophily, and ${\cal R}\ge 8$ indicating high structural heterophily. \\
\begin{table*}[ht]

\caption{Node-classification accuracies. Top three models are coloured by \textcolor{red}{ \textbf{First}}, \textcolor{blue}{ \textbf{Second}}, \textcolor{violet}{ \textbf{Third}}.}
\label{tab:results}
\begin{center}
\begin{small}
\begin{sc}
\resizebox{\textwidth}{!}{
\begin{tabular}{l c c c c c c c c c} 

                \toprule
                                \phantom{} & \textbf{Texas} & \textbf{Wisconsin} & \textbf{Cornell} & \textbf{Actor} & \textbf{Squirrel} & \textbf{Chameleon} & \textbf{Citeseer} & \textbf{Pubmed} & \textbf{Cora} \\
                \midrule
                                 Hom level & 0.11 & 0.21 & 0.30 & 0.22 & 0.22 & 0.23 & 0.74 & 0.80 & 0.81 \\ 
                                 ${\cal R}$ & \textbf{18.37} & \textbf{6.90} & \textbf{6.03} & \textbf{209.58} & \textbf{20.62} & \textbf{8.30} & \textbf{5.78} & \textbf{7.64} & \textbf{7.36}
                                 \\
                                \centering \# Nodes     & 183 & 251 & 183 & 7,600 & 5,201 &  2,277 & 3,327 &  19,717 & 2,708 \\ 
                                \centering \# Edges     & 295 & 466 & 280 & 26,752 & 198,493 & 31,421 & 4,676 & 44,324 & 5,278 \\ 
                                \centering \# Classes   & 5 & 5 & 5 & 5 & 5 & 5 & 7 & 3 & 6 \\ 
\midrule
                GGCN~\citep{GGCN} & $84.86\pm{4.55}$ & $86.86\pm{3.29}$ & ${85.68\pm{6.63}}$ & $\textcolor{blue}{37.54\pm{1.56}}$ & $55.17\pm{1.58}$ & $\textcolor{violet}{77.14\pm{1.84}}$ & $77.14\pm{1.45}$ & $89.15\pm{0.37}$ & $87.95\pm{1.05}$\\
                GPRGNN~\citep{GPRGNN} & $78.38\pm{4.36}$ & $82.94\pm{4.21}$ & $80.27\pm{8.11}$ & $34.63\pm{1.22}$ & $31.61\pm{1.24}$ & $46.58\pm{1.71}$ & $77.13\pm{1.67}$ & $87.54\pm{0.38}$  & $87.95\pm{1.18}$\\
                H2GCN~\citep{Beyond2020} & $84.86\pm{7.23}$ & $87.65\pm{4.89}$ & $82.70\pm{5.28}$ & $35.70\pm{1.00}$ & $36.48\pm{1.86}$ & $60.11\pm{2.15}$ & $77.11\pm{1.57}$ & $89.49\pm{0.38}$ & $87.87\pm{1.20}$\\
                GCNII~\citep{GCNII} & $77.57\pm{3.83}$ & $80.39\pm{3.40}$ & $77.86\pm{3.79}$ & $\textcolor{violet}{37.44\pm{1.30}}$ & $38.47\pm{1.58}$ & $63.86\pm{3.04}$ & $77.33\pm{1.48}$ & $\textcolor{blue}{90.15\pm{0.43}}$ & $88.37\pm{1.25}$\\
                Geom-GCN~\citep{GEOM-GCN} & $66.76\pm{2.72}$ & $64.51\pm{3.66}$ & $60.54\pm{3.67}$ & $31.59\pm{1.15}$ & $38.15\pm{0.92}$ & $60.00\pm{2.81}$ & $\textcolor{red}{78.02\pm{1.15}}$ & $89.95\pm{0.47}$ & $85.35\pm{1.57}$\\
                PairNorm~\citep{PairNorm} & $60.27\pm{4.34}$ & $48.43\pm{6.14}$ & $58.92\pm{3.15}$ & $27.40\pm{1.24}$ & $50.44\pm{2.04}$ & $62.74\pm{2.82}$ & $73.59\pm{1.47}$ & $87.53\pm{0.44}$ & $85.79\pm{1.01}$\\
                GraphSAGE~\citep{hamilton2017inductive} & $82.43\pm{6.14}$ & $81.18\pm{5.56}$ & $75.95\pm{5.01}$ & $34.23\pm{0.99}$ & $41.61\pm{0.74}$ & $58.73\pm{1.68}$ & $76.04\pm{1.30}$ & $88.45\pm{0.50}$ & $86.90\pm{1.04}$\\
                GCN~\citep{kipf2017semi} & $55.14\pm{5.16}$ & $51.76\pm{3.06}$ & $60.54\pm{5.30}$ & $27.32\pm{1.10}$ & $53.43\pm{2.01}$ & $64.82\pm{2.24}$ & $76.50\pm{1.36}$ & $88.42\pm{0.50}$ & $86.98\pm{1.27}$\\
                GAT\citep{velickovic2018gat} & $52.16\pm{6.63}$ & $49.41\pm{4.09}$ & $61.89\pm{5.05}$ & $27.44\pm{0.89}$ & $40.72\pm{1.55}$ & $60.26\pm{2.50}$ & $76.55\pm{1.23}$ & $87.30\pm{1.10}$ & $86.33\pm{0.48}$\\
                MLP & $80.81\pm{4.75}$ & $85.29\pm{6.40}$ & $81.89\pm{6.40}$ & $36.53\pm{0.70}$ & $28.77\pm{1.56}$ & $46.21\pm{2.99}$ & $74.02\pm{1.90}$ & $75.69\pm{2.00}$ & $87.16\pm{0.37}$\\
                CGNN\citep{CGNN} & $71.35\pm{4.05}$ & $74.31\pm{7.26}$ & $66.22\pm{7.69}$ & $35.95\pm{0.86}$ & $29.24\pm{1.09}$ & $46.89\pm{1.66}$ & $76.91\pm{1.81}$ & $87.70\pm{0.49}$ & $87.10\pm{1.35}$\\
                MixHop~\citep{mixhop19} & $77.84\pm{7.73}$ & $75.88\pm{4.90}$ & $73.51\pm{6.34}$ & $32.22\pm{2.34}$ & $43.80\pm{1.48}$ & $60.50\pm{2.53}$ & $76.26\pm{1.33}$ & $85.31\pm{0.61}$ & $87.61\pm{0.85}$\\
                FSGNN(8-hop)~\citep{FSGNN21} & $87.30\pm{5.29}$ & $87.84\pm{3.37}$ & $\textcolor{red}{87.84\pm{6.19}}$ & $35.75\pm{0.96}$ & $\textcolor{red}{74.10\pm{1.89}}$ & $\textcolor{blue}{78.27\pm{1.28}}$ & $\textcolor{violet}{77.40\pm{1.90}}$ & $77.40\pm{1.93}$ & $87.93\pm{1.00}$\\
                GRAFF~\citep{GRAFF} & $\textcolor{violet}{88.38\pm{4.53}}$ & $\textcolor{violet}{88.83\pm{3.29}}$ & $84.05\pm{6.10}$ & $37.11\pm{1.08}$ & $58.72\pm{0.84}$ & $71.08\pm{1.75}$ & $77.30\pm{1.85}$ & $\textcolor{violet}{90.04\pm{0.41}}$ & ${88.01\pm{1.03}}$\\
                LINKX~\citep{LINKX21} & $74.60\pm{8.37}$ & $75.49\pm{5.72}$ & $77.84\pm{5.81}$ & $36.10\pm{1.55}$ & ${61.81\pm{1.80}}$ & $68.42\pm{1.38}$ & $73.19\pm{0.99}$ & $87.86\pm{0.77}$ & $84.64\pm{1.13}$\\

                ACMII-GCN++~\citep{ACM} & $\textcolor{blue}{88.38\pm{3.43}}$ & $\textcolor{blue}{88.43\pm{3.66}}$ & ${86.49\pm{6.73}}$ & $37.09\pm{1.32}$ & $\textcolor{violet}{67.40\pm{2.21}}$ & $74.76\pm{2.20}$ & $77.12\pm{1.58}$ & $89.71\pm{0.48}$ & $\textcolor{violet}{88.25\pm{0.96}}$\\
                Ordered GNN~\citep{song2023ordered} & $86.22\pm{4.12}$ & $88.04\pm{3.63}$ & $\textcolor{violet}{87.03\pm{4.73}}$ & $\textcolor{red}{37.99\pm{1.00}}$ & $62.44\pm{1.96}$ & $72.28\pm{2.29}$ & $77.31\pm{1.73}$ & $\textcolor{red}{90.15\pm{0.38}}$ & $\textcolor{blue}{88.37\pm{0.75}}$
                \\
                ASGC~\citep{chanpuriya2022simplified} & $85.14\pm{3.06}$ & $86.06\pm{3.75}$ & $86.22\pm{3.58}$ & $36.33\pm{0.79}$ & $58.38\pm{1.08}$ & $73.16\pm{1.07}$ & $66.86\pm{0.86}$ & $78.72\pm{0.88}$ & $77.52\pm{1.61}$
                \\
\midrule
                \textbf{DJ-GNN} & $\textcolor{red}{92.43\pm{3.15}}$ & $\textcolor{red}{92.54\pm{3.70}}$ & $\textcolor{blue}{87.03\pm{1.62}}$ & $36.93\pm{0.84}$ & $\textcolor{blue}{73.48\pm{1.59}}$ & $\textcolor{red}{80.48\pm{1.46}}$ & $\textcolor{blue}{77.50\pm{1.33}}$ & $89.19\pm{0.32}$ & $\textcolor{red}{88.43\pm{0.91}}$ \\
                \bottomrule
\end{tabular}
}
\end{sc}
\end{small}
\end{center}
\end{table*}

\textbf{Low Structural Heterophily}. For these datasets, we list their optimal number of jumps (hyperparameter $K$): \textsc{Wisconsin} ($K=5$), \textsc{Cornell} ($K=5$), \textsc{Citeseer} ($K=5$), \textsc{Pubmed} ($K=5$) and \textsc{Cora} ($K=5$)) we find that a few jumps are enough for achieving or improving the SOTA. However, not all jumps are equally important. For instance, \textsc{Wisconsin} and \textsc{Cornell} rely mostly on the first two jumps (see Figure{~\ref{fig:alfas}}), while the remaining datasets rely on the \emph{homophilic branch} (no jump). Actually, \textsc{Citeseer}, \textsc{Pubmed}, and \textsc{Cora} are the datasets with the smallest edge heterophily (\textsc{Hom Level}). In addition, we are the second best method only in \textsc{Cornell} ($87.03$ (ours) vs $87.84$ (\textsc{FSGNN(8-hop)}) and we are very competitive in \textsc{Citeseer} ($77.50$ (ours) vs $78.02$ (\textsc{Geom-GCN})). In the first case, \textsc{FSGNN(8-hop)} uses a fully supervised split (60\%/20\%/20\%) whereas we use the more severe semi-supervised split. For \textsc{Citeseer}, we note that the \textsc{Geom-GCN} method relies on the geometry of the latent space. In this regard, \textsc{Citeseer} is the dataset with the lowest structural heterophily (${\cal R}=5.78$), i.e. the geometry of the latent space is a fair representation of the topology of the graph. As a result, adding jumps may complicate that geometry: actually, the most important branches are $\mathbf{J}^0$ and the \emph{homophilic branch} (no jump). Finally, in \textsc{Pubmed}, our method is slightly improved by \textsc{Ordered GNN} ($89.19$ vs $90.15$). \\

\textbf{High Structural Heterophily}. For these datasets~\citep{penn94}~\citep{ogbn}, we also list their optimal number of jumps (hyperparameter $K$): \textsc{Texas} ($K=20$), \textsc{Squirrel} ($K=8$), \textsc{Chameleon} ($K=12$) and \textsc{Actor} ($K=3$). Our best result is for \textsc{Texas} (${\cal R}=18.37$), where we significantly improve the SOTA ($92.43$ (ours) vs $88.38$ (\textsc{ACMII-GCN++}, which is a multi-channel GCN with adaptive channel mixing)). In \textsc{Squirrel}, we are slightly outperformed by \textsc{FSGNN(8-hop)} ($73.48$ vs $74.10$) due, again to their use of a full supervised split (60\%/20\%/20\%). We are very competitive in this dataset because the \textsc{Squirrel} graph is very dense and we only need $K=8$ jumps to achieve good results. We are also the best model in \textsc{Chameleon} (whose structural heterophily is the smallest in this set): we obtain $80.48$ vs the second-best model \textsc{FSGNN(8-hop)}  that achieves $78.27$ (again with a split of 60\%/20\%/20\%).

\textbf{Parallel (Shallow) vs Sequential (Deep)}. Our method is Parallel (multi-branch shallow GNN) and its performance is the best or it is very competitive in small-medium datasets. There is one exception. In the \textsc{Pubmed} dataset, where we obtain $89.15$, we are slightly outperformed by \textsc{Ordered GNN} with only $5$ layers ($90.15$) because this dataset is very homophilic. This also happens with \textsc{GCNII} which explores $2$ to $2^6$ layers. We can conclude that deep methods have a good performance in homophilic datasets but such a performance decays significantly in heterophilic ones. 
\begin{table*}[ht]
\caption{Node-classification accuracies in large graphs. Top three models are coloured by \textcolor{red}{ \textbf{First}}, \textcolor{blue}{ \textbf{Second}}, \textcolor{violet}{ \textbf{Third}}.}
\label{tab:results_large}
\begin{center}
\begin{small}
\begin{sc}
\resizebox{0.5\textwidth}{!}{
\begin{tabular}{l c c c} 

                \toprule
                                \phantom{} & \textbf{Penn94} & \textbf{arXiv-year} & \textbf{ogbn-arXiv}  \\
                \midrule
                                 Hom level & \textbf{0.47} & \textbf{0.21} & \textbf{0.66} \\ 
                                \centering \# Nodes     & 41,554 & 169,343 & 169,343  \\ 
                                \centering \# Edges     &  1,362,229 & 1,166,243 & 1,166,243 \\ 
                                \centering \# Classes   & 5 & 5 & 40  \\ 
\midrule
                MLP & $73.61 \pm{0.40}$ & $36.70\pm{0.21}$ & $55.91\pm{0.15}$
                \\
                GCN & $82.47\pm{0.27}$ & $46.02\pm{0.26}$ & $\textcolor{violet}{59.61\pm{0.23}}$
                \\
                GAT & $81.53\pm{0.55}$ & $46.05\pm{0.51}$ & $\textcolor{blue}{60.27\pm{0.21}}$
                \\
                MixHop & $\textcolor{violet}{83.47\pm{0.71}}$ & $\textcolor{blue}{51.81\pm{0.17}}$ & OOM
                \\
                LINKX & $\textcolor{blue}{84.71\pm{0.52}}$ & $\textcolor{red}{56.00\pm{1.34}}$ & $55.31\pm{0.81}$
                \\
\midrule    
                \textbf{DJ-GNN} & $\textcolor{red}{84.84\pm{0.34}}$ & $\textcolor{violet}{49.21\pm{0.20}}$ & $\textcolor{red}{63.23\pm{0.12}}$
                \\
                \bottomrule
\end{tabular}
}
\end{sc}
\end{small}
\end{center}
\end{table*}
\\
We have also tested our model in \textbf{Very Large Graphs} (see Table~\ref{tab:results_large}). In this regard, we note that the memory requirements of our method $-$ $O(n^2)$, where $n$ is the number of nodes $-$, force us to decouple the diffusion pump from the jump exploration. We first learn the matrix of pairwise diffusion distances in an unsupervised way. Later, we use it in a static way to minimize the classification loss. Despite that limitation, we obtain a very competitive performance both for \textsc{Penn94} ($84.84$ vs $84.71$ with \textsc{LINKX}) and  \textsc{ogbn-arXiv} ($63.23$ vs $60.27$ with \textsc{GAT}). However, our performance decreases in \textsc{arXiv-year} which is more heterophilic than the others: memory limitations force us to use only $K=3$ hops for the three datasets. 

Finally, we extend our experimental results in \emph{Appendix B: SBM Analysis} and \emph{Appendix C: Experimental and Computational Details}.

\section{Conclusions and Future Work}\label{sec:6}
In this paper, we propose \textsc{Diffusion-Jump GNNs}, a multi-branch GNN architecture that addresses the heterophily issue from a structural perspective. Firstly, we define node-classification in terms of a Dirichlet problem. This allows us to define a new measure of heterophily: structural heterophily. Having this measure in mind, we formulate a loss function that governs the interplay between the two main components of our architecture: the \textbf{diffusion pump} (which generates diffusion distances) and the \textbf{parallel jumps} (which drive the exploration of links between nodes with similar labels). The most important contribution of our model is that the diffusion distances, and consequently the jumps and the structural filters derived from them, are fully learnable. Our experiments show that our model outperforms the SOTA or it is very competitive. Finally, our future work includes: a) scalability, in terms of memory and b) automatic jump selection. 

\textbf{Broader Impact.} This research contributes to improving the reliability of graph learners in the heterophilic regime. Besides the well-known social benefits concerning the detection of malicious nodes in social networks, we consider the application of our learnable structural filters to fairness.


\medskip

{
\small

}
\newpage
\section{Appendix}\label{appendix}
\subsection{Appendix A: Formal Results with Practical Implications}
For the sake of clarity, in this appendix we develop the key concepts of the theorems stated in the paper instead of providing detailed proofs. Our emphasis here is on the practical implications of each result. For more details, we refer the reader to the cited papers. 

\textbf{The Trace Ratio Problem.} 
$\text{Min}\; \mathbf{U}^T\triangle\mathbf{U}$ is achieved by $\mathbf{V}_{n\times p}$ whose $p$ columns are given by the eigenvectors of the $p$ smallest eigenvalues $\lambda_1\le \lambda_2\le \ldots \le \lambda_p$ of the Laplacian $\triangle$. Then, $\text{Tr}[\mathbf{V}^T\triangle\mathbf{V}] = \lambda_2+\ldots + \lambda_p$, when the graph $G=(V,E)$ with adjacency matrix $\mathbf{A}$ is connected. However, $\mathbf{V}$ does not necessarily minimize  $\text{Tr}[\mathbf{V}^T\mathbf{D}\mathbf{V}]$. As a result, we have that 
\begin{equation}\label{eq:traceratAp1}
    \rho^{\ast}\defeq \text{Min}_{\mathbf{U}^T\mathbf{U}=\mathbf{I}}\frac{\text{Tr}[\mathbf{U}^T\triangle\mathbf{U}]}{\text{Tr}[\mathbf{U}^T\mathbf{D}\mathbf{U}]}
    \leq 
    \frac{\text{Tr}[\mathbf{V}^T\triangle\mathbf{V}]}{\text{Tr}[\mathbf{V}^T\mathbf{D}\mathbf{V}]}
    \leq
    \frac{\lambda_2+ \ldots + \lambda_p}{d_1 + \ldots + d_p}\;,
\end{equation}
where $d_1\le d_2\le \ldots \le d_n$ are the sorted degrees. As a result, we have the following bounds: 
\begin{equation}\label{eq:traceratAp2}
 \frac{\lambda_2+ \ldots + \lambda_p}{d_{p+1} + \ldots + d_n}
\leq
 \rho^{\ast} 
 \leq
\frac{\lambda_2+ \ldots + \lambda_p}{d_1 + \ldots + d_p}\;.
\end{equation}
The definition of $\rho^{\ast}$ plays a key role in the original trace-ratio optimization. Following~\citep{TraceRatio12}, such a problem is formulated in scalar terms, i.e. in terms of finding
\begin{equation}\label{eq:traceratAp3}
    \rho^{\ast} = \arg\min_{\mathbf{U}^T\mathbf{U}=\mathbf{I}} f(\rho)\defeq \text{Tr}[\mathbf{U}^T\triangle\mathbf{U}]-\rho\text{Tr}[\mathbf{U}^T\mathbf{D}\mathbf{U}]\;.
\end{equation}
Actually, for $\rho^{\ast}$ we have have that 
\begin{equation}\label{eq:traceratAp4}
    \text{Min}_{\mathbf{U}^T\mathbf{U}=\mathbf{I}} \text{Tr}[\mathbf{U}^T(\triangle-\rho^{\ast}\mathbf{D})\mathbf{U}] = 0\;.
\end{equation}
Therefore, the trace-ratio problem can be solved by alternating two updating steps:
\begin{itemize}
\item[$\mathbf{U}$] : Given $\rho$, apply the Lanczos method to obtain the $p$ largest eigenvalues of the transition matrix $\mathbf{P}-\rho\mathbf{D}$ (the smallest of $\triangle
 -\rho\mathbf{D}$) and their associated eigenvectors $\mathbf{U}$.
\item[$\rho$] :  Given the current eigenvectors $\mathbf{U}$,  update $\rho = \frac{\text{Tr}[\mathbf{U}^T\triangle\mathbf{U}]}{\text{Tr}[\mathbf{U}^T\mathbf{D}\mathbf{U}]}$\:
\end{itemize}
In the above process, the update of $\mathbf{U}$ ensures the orthogonality constraint. 
 
\textbf{The Trace Ratio and SGD.} 
However, solving the trace-ratio problem through gradient descent drives us to a different solution from the eigenvectors of $\triangle
 -\rho^{\ast}\mathbf{D}$. For instance, consider the Dirichlet loss ${\cal L}_D = \frac{\text{Tr}[\mathbf{U}^T\triangle\mathbf{U}]}{\text{Tr}[\mathbf{U}^T\mathbf{D}\mathbf{U}]}$. Then, its gradient (supposing that the orthogonality is enforced by a complementary loss) is given by: 
\begin{equation}\label{eq:traceratSGD1}
  {\nabla {\cal L}_D} \defeq  \frac{2\triangle\mathbf{U}-2\rho\mathbf{D}\mathbf{U}}{\text{Tr}[\mathbf{U}^T\mathbf{D}\mathbf{U}]}\;.
\end{equation}
Therefore, ${\nabla {\cal L}_D}=0$ implies $\triangle\mathbf{U}=\rho^{\ast}\mathbf{D}\mathbf{U}$, where $\rho^{\ast}\rightarrow 0$ is the asymptotic value of the trace ratio. As a result, we have that the optimal $\mathbf{U}$ satisfy $\tilde{\triangle}\mathbf{U}=\rho^{\ast}\mathbf{U}$, i.e. the gradient descent converges to the (orthonormal) functions of  the normalized Laplacian $\tilde{\triangle}$ associated with the value $\rho^{\ast}$. However, as $\rho^{\ast}$ is not necessarily an eigenvalue of $\tilde{\triangle}$, but it is close to the Fiedler value $\lambda_2$, we denote $\mathbf{U}$ as a \textbf{Fiedler environment}. It is an environment since the $p$ columns $\mathbf{u}_{:j}$ are mutually orthonormal and close to the Fiedler vector $\phi_2$ insofar their Dirichlet energies $\rho^{\ast}_j = {\mathbf{u}_{:j}}^T\tilde{\triangle}\mathbf{u}_{:j}$ satisfy $|\lambda_2-\rho^{\ast}_j|<\epsilon$ with $\epsilon\rightarrow 0$ (\textbf{Theorem}~\ref{th:1}).

In our experiments, we have chosen the trace-ratio formulation because:
\begin{itemize}
    \item [\textbf{a)}] It leads an \textbf{implicit normalization} of the gradient ${\nabla {\cal L}_D}$, namely $\text{Tr}[\mathbf{U}^T\mathbf{D}\mathbf{U}]$.  
    \item [\textbf{b)}] The \textbf{gradient is more structured} when we apply the constrain $\mathbf{U}=f_{\theta}(\mathbf{A})$, where $\mathbf{A}$ is the adjacency matrix. 
\end{itemize}

Regarding \textbf{a)}, our implicit normalization alleviates the problem of landing in local minima due to the orthonormalization constraint (that we also enforce in the global loss). As noted~\citep{SGDSIAM99},  constraints of the form $\mathbf{U}^T\mathbf{U}=\mathbf{I}$ define a Riemannian manifold and the trace problem s.t. them is not geodesically convex. In~\citep{SGDeigen18} this is addressed by introducing a Riemannian gradient and retraction normalization. 

However, our main gain in performance is achieved when we address \textbf{b)} via the joint effect of normalization and $\mathbf{U}=f_{\theta}(\mathbf{A})$. In our preliminary experiments, we compared the gradient when applying the constraint $\mathbf{U}=f_{\theta}(\mathbf{A})$ vs the one when doing only $\mathbf{U}=f_{\theta}(\mathbf{I})$. Discarding the biases, and the non-linearities in both cases, we have $\mathbf{U}=\mathbf{AW}$ vs $\mathbf{U}=\mathbf{W}$. For simplicity, we consider the gradient wrt a single column, i.e. we analyze $\mathbf{u}=\mathbf{Aw}$ vs $\mathbf{u}=\mathbf{w}$
\begin{equation}\label{eq:traceratSGD2}
  {\nabla {\cal L}'_{D,\theta}} \defeq  \frac{2(\triangle-\rho\mathbf{D})(\mathbf{Aw})}{\text{Tr}[(\mathbf{Aw})^T\mathbf{D}(\mathbf{Aw})]}\;\;\; \text{vs}\;\;\;
  {\nabla {\cal L}_{D,\theta}} \defeq  \frac{2(\triangle-\rho\mathbf{D})\mathbf{w}}{\text{Tr}[\mathbf{w}^T\mathbf{D}\mathbf{w}]}
  \;.
\end{equation}
Given a random initialization of $\mathbf{w}$, this vector plays the role of a random projector of the rows in $\mathbf{A}$. Following, the Johnson-Lindenstrauss Lemma,  $\hat{\mathbf{w}}=\mathbf{Aw}$ tends to replicate the structure of the adjacency. Actually, if the entries of $\mathbf{w}_i\sim {\cal N}(0,1)$ then, those of the projection satisfy $\hat{\mathbf{w}}_i\sim {\cal N}(0,d_i^2)$, where $d_i$ is the degree of node $i$. As a result, if we have $c$ well-defined communities in the graph $G=(V,E)$ with adjacency matrix $\mathbf{Aw}$, then the projection $\hat{\mathbf{w}}$ is near piecewise constant (actually the norm of the $i-$th row is preserved:  $\|\mathbf{A}_{i:}\|\approx \|\hat{\mathbf{\mathbf{w}}}\|$). As a result, the projection $\hat{\mathbf{w}}$ is more structured than  $\mathbf{w}$ and this is propagated and even amplified during gradient descent. In addition, the normalization of ${\nabla {\cal L}'_{D,\theta}}$ is stronger than that of ${\nabla {\cal L}_{D,\theta}}$. 

Overall, when evaluating the performance in \textsc{Squirrel} and \textsc{Chameleon} using only $\mathbf{U}=f_{\theta}(\mathbf{I})$ (i.e. using ${\nabla {\cal L}_{D,\theta}}$) we only obtain $41.38\pm{2.98}$ and $58.48\pm{4.69}$. However, using $\mathbf{U}=f_{\theta}(\mathbf{A})$ (gradient 
${\nabla {\cal L}'_{D,\theta}}$) leads to $73.48\pm{1.59}$ and $80.48\pm{1.46}$ respectively. 

Finally, a detailed impact of the two above formulations in the variance of the SGD as in~\citep{SGD19} is beyond the scope of this paper.  

\textbf{Assymptotic Diffusions.}  As explained above, optimizing the Dirichlet loss leads to Fiedler environments, i.e. the rows $\mathbf{U}_{i:}$ contain the $p$ nearly orthogonal eigenvectors with eigenvalues $\gamma_1=1>\gamma_2\ge\ldots\ge \gamma_p$. Following~\citep{difmaps05}, the \emph{diffusion distance} $D_j(i,j)$ at time $t$ between two nodes $i$ and $j$ is defined spectrally as: 
\begin{equation}\label{eq:diffdist}
D^2_t(i,j)\defeq \sum_k\frac{1}{\pi(k)}\left(p(k,t|i) - p(k,t|j)\right)^2
          = \Gamma^{2t}\|\mathbf{U}^{\ast}_{i:}-\mathbf{U}^{\ast}_{j:}\|^2\;,
\end{equation}

where $\pi(k)\defeq d_k/vol(G)$ are the components from the stationary probability distribution $\lim_{t\rightarrow\infty}p(j,t|i)={\mathbf \pi}$, i.e. the eigenvector $\mathbf{U}^{\ast}_{:1}$ corresponding to $\gamma_1=1$. In the above equation, $\mathbf{U}^{\ast}$ denote the true eigenvectors of the transition matrix $\mathbf{P}$, and $\Gamma\defeq\text{diag}(\gamma_1,\gamma_2,\ldots,\gamma_p)$ is the diagonal matrix with the corresponding eigenvalues. Thus, Eq.~\ref{eq:diffdist} can be explained in the following terms: 
\begin{itemize}
    \item [\textbf{a)}] $D^2_t(i,j)$ compare the probabilities that two random walks (one starting in $i$ and another one in $j$) reach any other node $k$ in time $t$.
    \item [\textbf{b)}] The spectral interpretation relies on the spectral theorem applied to the transition matrix $\mathbf{P}=\mathbf{U}^{\ast}\Lambda{\mathbf{U}^{\ast}}^T$. As a result, $\mathbf{P}^t=\mathbf{U}^{\ast}{\Gamma}^t{\mathbf{U}^{\ast}}^T=\sum_{r=1}^n\gamma_r^{t}\mathbf{U}^{\ast}_{:r}{\mathbf{U}^{\ast}_{:r}}^T$, with $n=|V|$. 
\end{itemize}
However, since determining what is the correct diffusion time is very hard (it is usually a hyperparameter in some GNNs), we are interested in the asymptotic diffusion distance $D^2_{t\rightarrow\infty}$. Qiu and Hancock~\citep{Qiu07CTembedding} determined that 
\begin{equation}\label{eq:edwin}
    \sum_{t=0}^{\infty}D^2_{t}(i,j)=\sum_{r=2}^{n}\frac{1}{1-\gamma_r}\left(\mathbf{U}^{\ast}_{ri}-\mathbf{U}^{\ast}_{rj}\right)^2
\end{equation}
i.e. eigenvalues $\{1-\gamma_r\}$ of the normalized Laplacian $\tilde{\triangle}$ are used instead of those of $\mathbf{P}$. Actually, the right side of the above equation is the well-known \emph{commute times}~\citep{Chandra89} distance $\text{CT}(i,j)$. Note that such a distance is dominated by the Fiedler value and vector: $\lambda_2=(1-\gamma_2)$ and $\mathbf{U}^{\ast}_{:2}$, respectively. This fact simplifies the interpretation of our approximate diffusion distance as follows:
\begin{itemize}
    \item [\textbf{a)}] Our approximated eigenvectors, contained in the $p$ columns of $\mathbf{U}$ have eigenvalues (Dirichlet energies) $\rho^{\ast}_r$ close to $\rho^{\ast}$ (the optimal trace ratio achieved by the Dirichlet loss). 
    \item [\textbf{b)}] Theorethically, we have that the smallest $\rho^{\ast}_{r_\text{min}}$ satisfies $\lambda_2\le \rho^{\ast}_{r_\text{min}}$. Therefore, if we order $\rho^{\ast}_{r}$ in ascending order, then we obtain  
    \begin{equation}\label{eq:edwin2}
       \sum_{t=0}^{\infty}D^2_{t}(i,j)\approx \sum_{r=1}^{p}\frac{1}{\rho^{\ast}_r}\left(\mathbf{U}_{ri}-\mathbf{U}_{rj}\right)^2\approx \text{CT}(i,j)\;.
    \end{equation}
    \item [\textbf{c)}] However, in the heterophilic regime (where the labels break the structure) we usually have $\lambda_2\ll \rho^{\ast}_{r_\text{min}}$. See for instance the Fiedler environments obtained for SBMs in Figure~\ref{fig:sbm} and the discussion below (\emph{Appendix B}). As a result, in practice we have
    \begin{equation}\label{eq:propct}
        d^{t\rightarrow \infty}(i,j)\defeq \sum_{r=1}^{p}\left(\mathbf{U}_{ri}-\mathbf{U}_{rj}\right)^2 = \|\mathbf{U}_{i:}-\mathbf{U}_{j:}\|^2=\alpha\text{CT}(i,j)\;\; \text{where}\;\; \alpha\ll 1.
    \end{equation}
    This proofs \textbf{Corollary}~\ref{cor:1}.    
\end{itemize}

\textbf{Escape Probabilities.} Approximating commute times distances is very convenient for our jump-based analysis, since it is well known that the \emph{escape probability} is related to the commute times distance~\citep{doyle2000random}: $p_{esc}=\frac{1}{\text{CT}(i,j)}$. Escape probabilities are actually dependent on the spectral gap (approximated by the Fiedler value $\lambda_2$). This is illustrated in the very first Figure of this paper (Figure~\ref{fig:jump}), where a random walker tries to escape from the community $\bar{A}$. A classic result~\citep{pmlr-vR3-meila01a} shows that the probability that a random walk started in its asymptotic ($t\rightarrow\infty$) distribution $\pi$ is transitioning from $i\in \bar{A}$ to $j\in A$ in one step is $p_{esc}(\bar{A},A)=\frac{\text{cut}(\bar{A},A)}{\text{vol}(\bar{A})}$, where $\text{cut}(\bar{A},A)=\sum_{i\in \bar{A},j\in A}e_{ij}$ (in the Figure we have that $\text{cut}(\bar{A},A)=1$). 

Therefore, as $d^{t\rightarrow \infty}(i,j)=\alpha\text{CT}(i,j)$, with $\alpha\ll 1$ our jump-hierarchy is closer to that of the escape probability than choosing $\text{CT}(i,j)$ as asymptotic diffusion distance. In addition, we are sensitive to the spectral gap since the Fiedler environment contains approximations of the Fiedler vector, and the spectral gap is approximated in turn by the Dirichlet energy of the Fiedler vector.

\textbf{Clustering and Metastable States.} Minimizing the Dirichlet loss in conjunction with the classification loss (see Eq~\ref{eq:loss}) leads to a trade-off between two \emph{clustering} problems. On the one hand, we infer a piecewise-smooth latent space. On the other hand, we simultaneously try to preserve the structure of the input graph as much as possible. In both cases, we try to find metastable states. A metastable state is a concept borrowed from dynamical systems but basically, it is an equilibrium state in a random process (for instance the one defined by a random walk that tries to escape from a community in Figure~\ref{fig:jump}). Metastable states are also characterized by wells in potential functions $U(\mathbf{x})$, where $\mathbf{x}$ is a state and its probability is given by the Boltzmann distribution $p(\mathbf{x})=e^{-U(\mathbf{x})}$. 
Then, the characteristic relaxation processes and time scales of a given space are usually described by a Stochastic Diffusion Equation (SDE): 
\begin{equation}
    \dot{\mathbf{x}}(t) = - \nabla U(\mathbf{x}) + \sqrt{2}\dot{\mathbf{w}}(t)
\end{equation}
where $\mathbf{w}$ denotes Brownian motion. In the above equation, we have a \emph{drag} term (the gradient) that drives the process to a deep well, and a \emph{random} term (the Brownian motion) that allows us to escape from local minima. During this process, we find different time scales: fast scales while we are moving through a given well, and slow scales when we try to escape from it. For instance, escaping from the right community in Figure~\ref{fig:jump} takes a long time which depends on the difference between the potential at the well $U(\mathbf{x}_{min})$ and that of the saddle point $U(\mathbf{x}_{max})$~\citep{sclimitations06}. This time is in turn the inverse of the spectral gap, i.e. there is a spectral interpretation of the SDE. Such interpretation comes from the analysis of the Fokker-Planck equation:
\begin{equation}
   \partial_t p(\mathbf{x},t) = \nabla\cdot\left[\nabla p(\mathbf{x},t) + p(\mathbf{x},t)\nabla U(\mathbf{x})\right]\;,
\end{equation}
This equation leads to the pdf of the SDE and it has a spectral interpretation. More precisely, the eigenvectors of $\mathbf{P}$ converge to the eigenfunctions $\Psi(\mathbf{x})$ of the Fokker-Planck equation as follows~\citep{sclimitations06}\citep{NADLER2006113}:  
\begin{equation}
   \tilde{\nabla}\Psi(\mathbf{x})\defeq \triangle\Psi - \nabla\Psi\cdot\nabla U = -\mu\Psi(\mathbf{x})\;,
\end{equation}
where $\triangle=\nabla\cdot\nabla$ is the Laplacian and $\mu$ the eigenstates (eigenvalues). As a result, we may use the Fiedler vector to characterize the separation between two clusters. The steepest the Fiedler vector, the better the separation (\textbf{Theorem}~\ref{th:3}). Interestingly, the third eigenvector $\tilde{\nabla}$ may not work well as a state separator when we have different spatial scales, as we will see in \emph{Appendix B}.

\textbf{Dirichlet Label Propagation.} As we mentioned in the main paper, our Dirichlet formulation is inspired by classical graph-based semi-supervised methods. The work in~\citep{Zhou03} poses the problem of propagating known labels $\ell(B)$ to unknown nodes $u\in U$. Let $\mathbf{Y}$ be a $n\times c$ matrix where $\mathbf{Y}(i,j)=1$ means that node $i\in B$ has label $j$ and $\mathbf{Y}(i,j)=0$ otherwise. Then, the optimal label of each node $i$ is given by $\mathbf{y}(i)=\arg\max_{j\leq c}\mathbf{F}(i,j)$, where
\begin{equation}
    \mathbf{F}=\beta\left(\mathbf{I}-\alpha\mathbf{P}\right)^{-1}\;,
\end{equation}
being $\mathbf{P}$ the transition matrix and $\alpha + \beta = 1$. The $n\times c$ matrix $\mathbf{F}$ works as a basic node representation (not exactly a latent space) since each of is $c$ rows is stochastic. However, its construction exploits the powers of $\mathbf{P}$ as follows: 
\begin{equation}\label{eq:DirichletLocal}
    \mathbf{F}(t) = (\alpha\mathbf{P})^{t-1}\mathbf{Y} + (1-\alpha)\sum_{i=0}^{t-1}(\alpha\mathbf{P})^{i}\mathbf{Y}\;.
\end{equation}
and
\begin{equation}
    \mathbf{F} = \lim_{t\rightarrow\infty} \mathbf{F}(t)=(1-\alpha)(\mathbf{I}-\alpha\mathbf{P})^{-1}\;.
\end{equation}
In addition, we also noted that $\mathbf{F}$ is also the solution of the Dirichlet problem:
\begin{equation}
    \text{Min}\;\; {\cal L}_Q=\frac{1}{2}\left(\text{Tr}[\mathbf{F}^T\triangle \mathbf{F}] + \mu\sum_i\|\mathbf{F}_{i:} - \mathbf{Y}_{i:}\|^2\right)\;.
\end{equation}
where $\mu>0$ is a regularization parameter satisfying $\alpha=1/(1+\mu)$. The proof is obtained by setting the gradient to zero:
\begin{eqnarray}
    \nabla {{\cal L}_Q}_{\mathbf{F}} &=& \mathbf{F} - \mathbf{P}\mathbf{F} + \mu(\mathbf{F}-\mathbf{Y})\nonumber\\
    &=& \mathbf{F} - \frac{1}{1 + \mu}\mathbf{P}\mathbf{F} - \frac{\mu}{1 + \mu}\mathbf{Y}\nonumber\\
    &=& \mathbf{F} - \alpha\mathbf{P}\mathbf{F} - \beta\mathbf{Y} = 0\;,
\end{eqnarray}
which leads to Eq.~\ref{eq:DirichletLocal}.

Concerning the relationship of this formulation with absorbing random walks~\citep{doyle2000random}, 
the main idea is to extend the $n\times n$ transition matrix $\mathbf{P}$ so that:
\begin{itemize}
    \item [\textbf{a})] We include an upper block with the $p\times p$ identity matrix $\mathbf{I}$. This block represents the $p$ absorbing states, where $p=|\ell(B)|$. Then the $n\times p$ block $\mathbf{R}$ encodes the prior probabilities of reaching an absorbing state from a non-absorbing one. 
    \item [\textbf{b})] The absorbing probabilities are given by $\mathbf{B}\defeq (\mathbf{I}_{n\times n}-\mathbf{P})^{-1}\mathbf{R}$\;.
\end{itemize}
Finally, the random-walker version~\citep{Grady06} is quite similar to the above one, but reorganizes the Laplacian matrix (\textbf{Theorem}~\ref{th:2}).




\subsection{Appendix B: SBM Analysis}
The following experiment aims to illustrate the interplay between our novel measure of structural heterophily ${\cal R}$ and the extent of the spectral gap. We also show the Fiedler Environments and how they are influenced by the classification loss (labels). For each heterophilic regime, we show both the corresponding pairwise distance matrix (diffusion map) and the resulting homophiliation. 
\begin{figure*}[ht]
\begin{center}
\centerline{\includegraphics[width = \linewidth,height =7cm]{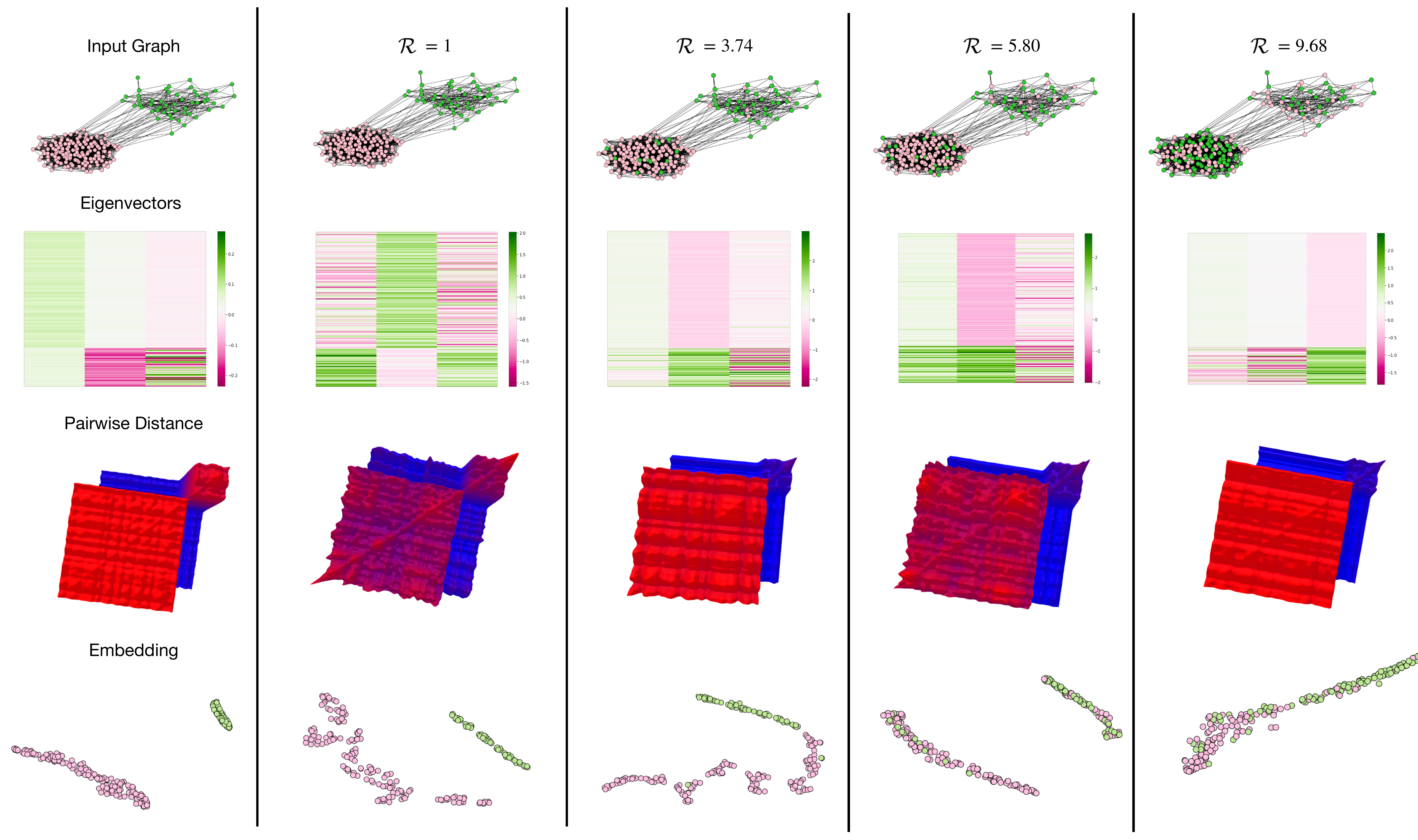}}
    \caption{Structural Heterophily in SBMs. Left: The original homophilic graph (First row), its $p=3$ eigenvectors (Second row), the pairwise distance matrix (Third row), and the resulting embedding. The remaining columns to the right have the same structure for increasing levels of structural heterophily. Note the evolution of the Fiedler Environments and the homophiliations. In all cases, we use $K=10$ jumps.}
    \label{fig:sbm}
\end{center}
\end{figure*} 
We have depicted in Figure~\ref{fig:sbm} the main ingredients of our approach as a means of illustrating some technical details introduced in \emph{Appendix A}. In particular, when analyzing SBM graphs under structural heterophily we observe several interesting phenomena.  

\textbf{Original vs Learned}. Instead of precalculating the eigenvectors, as in Directional GNNs~\citep{beaini2021directional}, we learn them. Our learned (approximate) eigenvectors are relatively close to the Fiedler vector (in terms of how they discriminate the two classes). This is what we call \emph{Fiedler Environments} but, in a semi-supervised setting, i.e. the learned eigenvectors are reactive both to the Dirichlet loss and to the classification loss. Despite being noisy, the vectors in the Fiedler Environments are able of partitioning a class when needed (especially for high values of ${\cal R}$). 
    
\textbf{Diffusion Map.}  Our pairwise distances are also reactive to semi-supervised classification. However, the Dirichlet loss tends to flatten the intra-communities distances as much as possible. Flattening is a mechanism to enforce intra-community diffusion in the homophilic regime. In the heterophilic regime, however, the diffusion map enforces exploration via high-order jumping (see lateral steps in the blue region and the loss of the red peak in the small community). 

\textbf{Embedding.} We can also see how the embedding is affected by structural heterophily. When we have a structural cluster with nodes of two classes, the respective embeddings are correctly separated, but the margin of this separation decreases as ${\cal R}$ increases. This can be seen in graphs that have high $\cal{R} \gg$ 1, where it is common to find subclusters of nodes that belong to the other classes.
\begin{figure*}[ht]
\begin{center}
\centerline{\includegraphics[width = \linewidth,height =7cm]{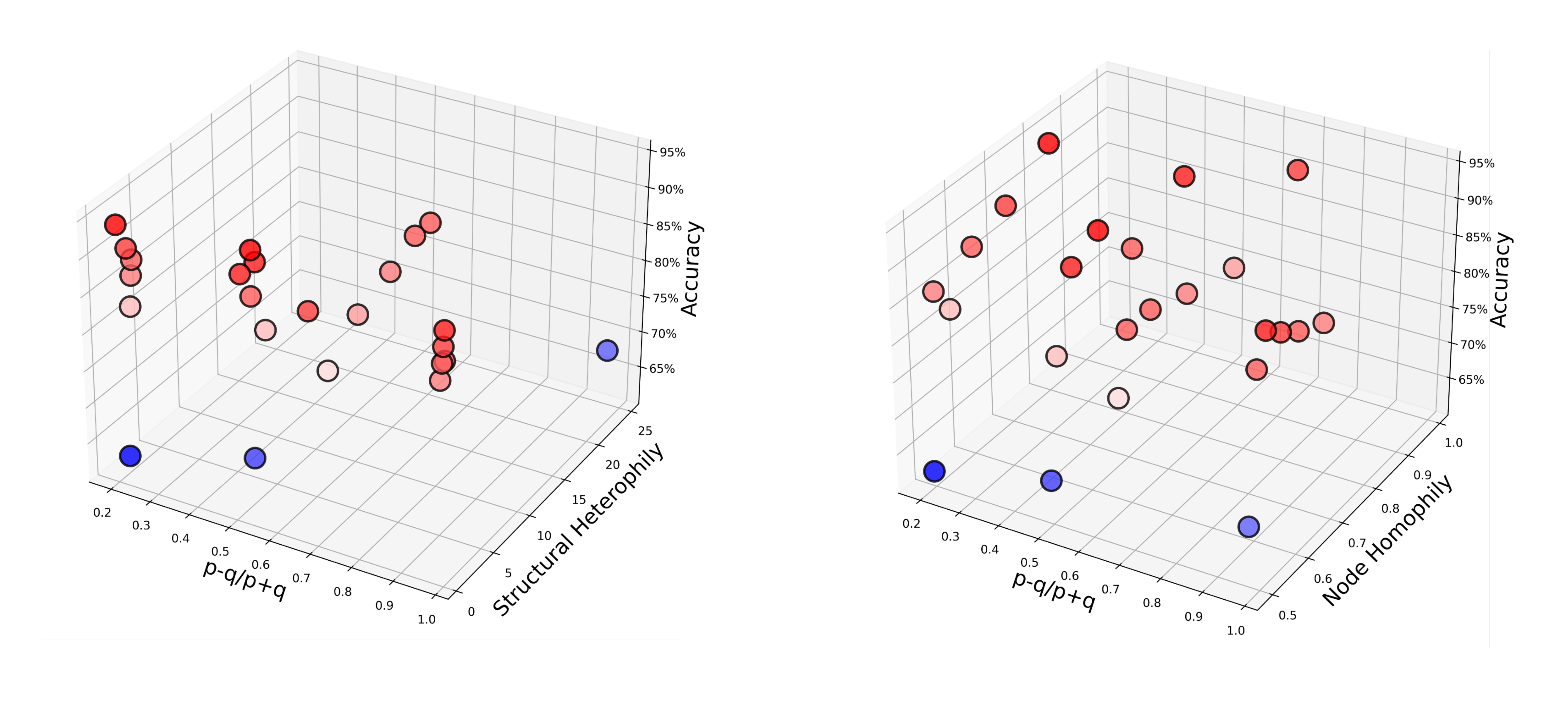}}
    \caption{Interplay between heterophily and the spectral gap $\frac{p-q}{p+q}$. Left: Results wrt structural heterophily. Right: Results wrt node homophily.}
    \label{fig:sh-explain}
\end{center}
\end{figure*} 

\textbf{Interplay between ${\cal R}$ and the gap.} Finally, we extend the experiments of~\citep{ASGC} by incorporating a third axis in addition to variate $p$ and $q$. This new axis is the structural heterophily. 
We proceed as follows. We generate four \emph{basic SBMs} attending to increasing spectral gaps: $\frac{p-q}{p+q} \in \{0.2, 0.5, 0.67, 0.98\}$. For each basic SBM we have generated six levels of increasing structural heterophily ${\cal R}$. In parallel, we also generate six levels of increasing node homophily as a means of complementing structural heterophily.

We show our results in Figure~\ref{fig:sh-explain}:
\begin{itemize}
    \item[\textbf{a)}] \textbf{Small Gaps help.} Our method is based on spectral clustering, which means that keeping the gap low factor is key. This helps our method to choose whether to jump outside the cluster looking for a node with the same label (Heterophilic regime) or to stay and only look around (homophilic regime). This common case is supported by our method without problems.
    \item[\textbf{b)}] \textbf{Low/Medium Structural Heterophily.} If the structure is quite correlated with the label and the spectral gap is not too high, our method is able of achieving good results even when the structure is noisy.
    \item[\textbf{c)}]  \textbf{Large Gaps lead to oversmoohing.} Our worst performance is achieved when the inter-class message passing is massive.   This leads to oversmothing due to the high connectivity of the graph. This high connectivity cannot be controlled by our pump (see the blue dots).
\end{itemize}
We have also performed the same experiment, but changing the structural heterophily measure to node homophily, in order to display the difference between both. Note that our measure fails when the spectral gap is large. This happens because the Dirichlet energy in a near-complete graph is minimal. This lack of structure leads ${\cal R}$ to consider that all the nodes are in the same cluster, i.e. is no heterophily).

\subsection{Appendix C: Experimental and Computational Details}
\begin{figure*}[ht]
\begin{center}
\centerline{\includegraphics[width = 0.6\linewidth,height =8cm]{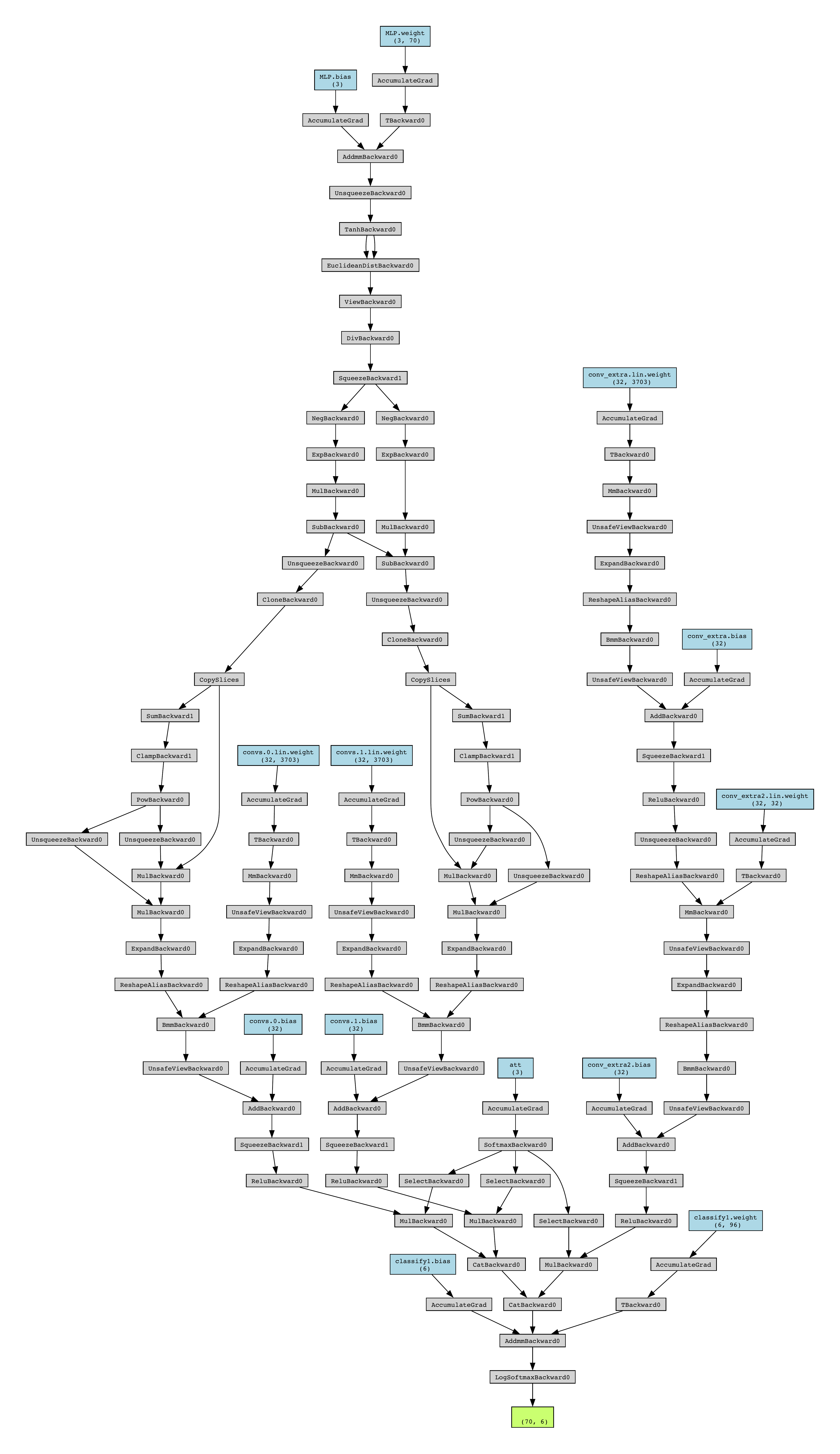}}
    \caption{The Computational Graph for $K=3$ jumps. All branches depend on the diffusion pump (top-left) except $HB$ (the Homophily Branch, top-right).}
    \label{fig:topk-explain}
\end{center}
\end{figure*}

In this section, we provide details about the datasets (see Table~\ref{tab:datasets-table}) and all the parameters and configurations of our experiments (see Table~\ref{tab:datasets-tuning} in order to better clarify the architecture and the results. \textsc{Diffusion-Jump GNNs} is implemented in PyTorch~\citep{pytorch}, using PyTorch Geometric~\citep{pyg} and ogbn datasets~\citep{ogbn}. For reproducibility, code, and instructions are available on our GitHub with all the selected configurations and logs. We have also included the computational (See Figure~\ref{fig:topk-explain}) in order to clarify the derivability of $topk$ in PyTorch.
\begin{table}[ht]
\caption{Statistics of the datasets used in our experiments.}
\label{tab:datasets-table}
\begin{center}
\begin{small}
\begin{sc}
\resizebox{0.5\textwidth}{!}{
\begin{tabular}{l c c c c} 

                \toprule
                                \textbf{Dataset} &  \textbf{Avg D} & \textbf{Density}  & \textbf{Node H} &\textbf{Class H}\\
                \midrule
                Texas & 1.77 & 0.0090 & 0.07 & 0.001  \\ 
                Wisconsin  & 2.05 &  0.0080 & 0.17 & 0.094 \\
                Cornell   &  1.62 & 0.0080 & 0.11 & 0.047\\
                Actor  & 3.94 & 0.0005 & 0.16 & 0.011\\
                
                Squirrel & 41.73 & 0.0080 & 0.09 & 0.025\\
                Chameleon  & 15.85 & 0.0070 & 0.10 & 0.062  \\
                Citeseer  & 2.73 & 0.0008  & 0.71 & 0.627   \\
                Pubmed  & 4.49 & 0.0002 &  0.79 & 0.664  \\
                Cora  & 3.89 & 0.0014 &  0.83 & 0.776 \\
                Penn94  & 3.89 & 0.0014 &  0.83 & 0.776 \\
                Ogbn-arXiv  & 7 & 0.0004 &  0.66 & 0.444 \\
                ArXiv-year & 7 & 0.0004 &  0.22 & 0.272 \\
                \bottomrule
\end{tabular}
}
\end{sc}
\end{small}
\end{center}
\end{table}

In the following Table~\ref{tab:datasets-tuning}, we include the hyperparameters that have yielded the best results during the experimentation phase. It is worth noting that the experiments were conducted using the same 10 random splits as in~\citep{pei2020geom}, training during 700 epochs and utilizing early stopping.

\begin{table}[ht]
\caption{Best hyperparameters for our datasets.}
\label{tab:datasets-tuning}
\begin{center}
\begin{small}
\begin{sc}
\resizebox{0.8\textwidth}{!}{
\begin{tabular}{l c c c c c} 

                \toprule
                                \textbf{Dataset} &  \textbf{hidden channels} & \textbf{dropout} &
                                \textbf{lr} &
                                \textbf{weight decay} & \textbf{k/\#Jumps}\\
                \midrule
                Texas      & 64  & 0.2  & 0.03  & 0.0005 & 20  \\ 
                Wisconsin  & 64  & 0.5  & 0.03  & 0.0005 & 5  \\
                Cornell    & 128 & 0.5  & 0.03  & 0.001  & 5  \\
                Actor      & 16  & 0.2  & 0.03  & 0.0001  & 3\\
                Squirrel   & 128 & 0.5  & 0.003 & 0.0005 & 8\\
                Chameleon  & 128 & 0.35 & 0.003 & 0.0005 & 12  \\
                Citeseer   & 128 & 0.5  & 0.003 & 0.0005 & 5 \\
                Pubmed     & 128 & 0.3  & 0.01  & 0.0005 & 3 \\
                Cora       & 128 & 0.5  & 0.002 & 0.0005 & 5\\
                Penn94     & 16  & 0.5  & 0.001 & 0.0001 & 3\\
                Ogbn-arXiv & 128 & 0.3  & 0.01  & 0.0005 & 3\\
                ArXiv-year & 128 & 0.2  & 0.003 & 0.0005 & 3\\
                \bottomrule
\end{tabular}
}
\end{sc}
\end{small}
\end{center}
\end{table}





\end{document}